\documentclass[10pt,journal,compsoc]{IEEEtran}
\usepackage{amsmath,amsfonts}
\usepackage{algorithmic}
\usepackage{array}
\usepackage{makecell}
\usepackage{textcomp}
\usepackage{stfloats}
\usepackage{url}
\usepackage{verbatim}
\usepackage{graphicx}
\usepackage[caption=false,font=footnotesize,labelfont=sf,textfont=sf]{subfig}
\usepackage[nocompress]{cite}
\usepackage{wrapfig}
\usepackage{multirow}
\usepackage[table,xcdraw]{xcolor}
\begin{document}

\title{GSCV++: Lightweight 3D Gaussian Splatting Sequence Compression via Standard Video Codecs}

\author{Qi Yang,~\IEEEmembership{Member,~IEEE,} Yingzhan Xu, Kai Zhang,~\IEEEmembership{Senior Member,~IEEE}, Li Zhang,~\IEEEmembership{Senior Member,~IEEE}
\thanks{Q. Yang, K. Zhang, and L. Zhang are from Bytedance Inc., San Diego (e-mail: \{chino.yang, zhangkai.video, lizhang.idm\}@bytedance.com) }
\thanks{Y. Xu is from Bytedance Inc., Beijing (e-mail: xuyingzhan@bytedance.com)}
}

\markboth{}{}


\IEEEtitleabstractindextext{

\begin{abstract} This paper presents GSCV++, a lightweight pipeline for 3D \textbf{G}aussian \textbf{S}platting (GS) sequences \textbf{C}ompression with standard \textbf{V}ideo codecs. To meet the requirements of low-complexity and hardware compatibility from the industry, GSCV++ jointly improves preprocessing, 3D-to-2D mapping, and practical video compression configuration of GS compression. First, for preprocessing, energy-adaptive inter-frame spherical harmonic (SH) alternative current (AC) dimensionality reduction (EA-SHDR) is proposed to produces a compact and stable temporal SH AC feature space. It can reduce the data volume of following mapping and practical compression, as well as less complexity while eliminating redundant information. Second, for 3D-to-2D mapping, prior-based Inter-MiniPLAS (P-MiniPLAS) is designed to generate smooth 2D feature maps through prior-based weighting factor inference and small scale remapping. Compared with vanilla Parallel Linear Assignment Sorting (PLAS) and Morton scan, P-MiniPLAS ensures the generated 2D maps are more suitable for the video codec while maintaining a low level of computational complexity.  Third, for practical compression, content-aware generalized compression (CA-GC) is developed to use a non-uniform bit-depth split for coordinate map generation, and a coding unit size selection is introduced for efficient compression complexity control, which fully exploits the performance of video codecs without involving normative changes. Experiments demonstrate that GSCV++ reports over 20\% in intra- and 60\% in inter-prediction rate-distortion gains compared with state-of-the-art (SOTA) methods. GSCV++ also achieves faster 2D map generation, encoding, and decoding than other SOTA methods using the same type of video codecs. Ablation studies verify the effectiveness and robustness of the proposed modules and provide insights for subsequent research. \end{abstract}

\begin{IEEEkeywords}
Gaussian splatting, compression, video codec
\end{IEEEkeywords}}\maketitle

\section{Introduction}
\IEEEPARstart{3}{D} Gaussian splatting (GS) \cite{vanillaGS} achieves state-of-the-art (SOTA) performance in 3D reconstruction, as well as an impressive trade-off between generation speed and rendering quality. Different from the neural radiance field (NeRF) \cite{mildenhall2021nerf} that embeds scene information into neural networks, GS represents 3D scenes using numerous primitives with multi-channel attributes, including geometric coordinates, color spherical harmonic (SH) direct current (DC), color SH alternating current (AC), opacity, scale, and rotation. The high-dimensional explicit representation imposes substantial memory and storage demands, leading to considerable attention to 3D GS compression in both academia and industry.

The Moving Picture Experts Group (MPEG) has considered two approaches to compress 3D GS: A-3DGS and I-3DGS \cite{m71640}. A-3DGS uses multiview images/videos as input and generates a compact representation through an optimized reconstruction process, as in HAC \cite{chen2024hac}, HAC++\cite{chen2025hac++}, and CompGS \cite{liu2024compgs}. I-3DGS takes generated 3D GS data as input, generates bitstreams, and reconstructs the symmetrical GS data, as in GGSC \cite{yang2024benchmark}, LGSCV \cite{LGSCV}, and HGSC \cite{huang2024hierarchical}. A-3DGS and I-3DGS differ in their input requirements, optimization scope, and computational resources. Generally, A-3DGS methods can report better rate-distortion (RD) results than I-3DGS \cite{xing20253dgs}, while also requiring higher encoding times and heavier computational resources.  Recently, a joint industry proposal \cite{m72430} motivates lightweight GS compression through the reuse of standard video and point cloud codecs, such as HEVC \cite{sullivan2012overview} and GPCC \cite{pcc-1}, to reduce deployment and development costs. Therefore, I-3DGS is regarded as a higher priority, i.e., fast track, in the current standardization stage, such as V-PCC AMD1 \cite{vpccAmd1}.

\begin{figure}[!t] \centering \includegraphics[width=0.5\textwidth]{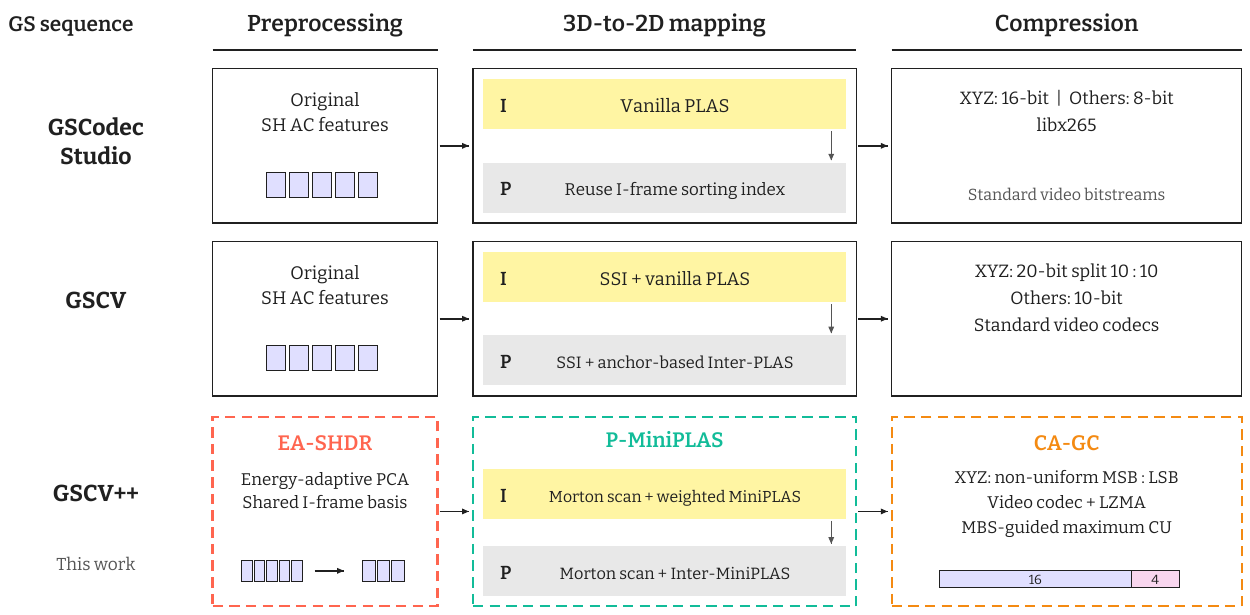} \caption{Comparison of video-based GS compression pipelines. GSCodec Studio, GSCV, and GSCV++ are contrasted across preprocessing, 3D-to-2D mapping, and compression.} \label{fig:motivation} \end{figure}

In this paper, we follow the I-3DGS track and focus on reusing standard video codecs to compress preconstructed GS sequences \cite{m72430}, which involve three main steps, as shown in Fig. \ref{fig:motivation}: preprocessing, 3D-to-2D mapping, and the compression module. First, given a 3D GS sample, a preprocessing step is applied to reduce the data volume by removing redundant information, such as pruning \cite{Etri-pruning, kaifa-pruning} and feature dimensionality reduction \cite{pca-wg4, etri-pca}. Second, 3D GS to 2D mapping is a key step for efficient GS coding via video codecs. Because standard video codecs are optimized for natural image content, their performance improves as the generated 2D maps better match the distribution of natural videos. Finally, the compression module implements the actual compression process, converting the mapped 2D maps into standard bitstreams.

The current MPEG video-based GS compression anchor, i.e., GSCodec Studio \cite{li2025gscodec}, uses Parallel Linear Assignment Sorting (PLAS) \cite{PLAS} to convert GS into smooth 2D maps, followed by FFMPEG libx265 to perform compression. It uses a bit depth (BD) of 16 for geometric coordinates and 8 BD for other attributes to generate quantized images/videos. Based on preliminary results, GSCodec Studio performs stably and effectively on the MPEG tracked dataset. However, there are three limitations of GSCodec Studio: 1) SH AC attributes occupy the majority of the data volume, and GSCodec Studio directly converts them into 2D maps and feeds them into the video codec, which can lead to a lower compression ratio and higher coding complexity.    2) The vanilla PLAS algorithm is a heuristic algorithm that progressively reorders the GS maps from an extremely large receptive field, which is relatively time-consuming (see section \ref{sec:preli_plas} for details) and difficult to converge to optimal results. For example, PLAS requires around 30 seconds on GS samples containing approximately half a million primitives.  3) For sequence compression, GSCodec Studio reuses the I-frame PLAS permutation for later frames and, therefore, assumes consistent primitive indices. If no tracked information is provided by the GS sequences, or if some preprocessing operators (e.g., pruning) destroy the tracked information, the P-frame 2D maps resemble noise-like images and incur poor inter-frame compression results, limiting the applicability of this strategy.


 The current study has developed methods to address the problems mentioned above. First, {for preprocessing}, some research has proven that SH AC illustrates high inter-channel redundancy \cite{yanghybridgs} for a single GS frame. However, how to reduce inter-frame SH AC redundancy while keeping stable rendering quality still requires further study. Second, {for 3D-to-2D mapping}, a more efficient and lightweight 2D map generation method is proposed by LGSCV \cite{LGSCV}, in which the two-stage Morton initialization and MiniPLAS are utilized. However, it also only considers the all-intra case. Third, for sequence compression, GSCV \cite{GSCV} proposes an inter-PLAS algorithm to build the track information between adjacent frames, which is more robust to different GS sequences. However, inter-PLAS significantly increases the computational complexity; a more efficient correspondence reconstruction is desired.

To develop a more effective and lightweight GS sequence compression framework, we propose GSCV++ to improve the RD and processing-time trade-off without supplied inter-frame tracked information.  First,  we design an energy-adaptive inter-frame SH AC dimensionality reduction (EA-SHDR) module. Based on principal component analysis (PCA), EA-SHDR selects an I-frame dimension using explained variance and reuses the basis and mean for the following frames, which can extend earlier all-intra dimensionality reduction \cite{LGSCV,pca-wg4} to a common and stable feature space within a GoP in an extremely low-complexity implementation.  

Second, a prior-based Inter-MiniPLAS refinement (P-MiniPLAS) is proposed to realize fast and smooth 2D map generation for inter-mode. P-MiniPLAS is a variation of vanilla PLAS, which consists of two main enhancements: First, vanilla PLAS assigns equal weighting factors to all the GS attributes during the reordering process, meaning that all selected features use the same scaling factor in the reordering objective. However, GS attributes have different importance in terms of rendering quality, as well as different sensitivities to lossy compression \cite{adcgs++}. Generally, for certain GS attributes, the smoother the 2D map is, the fewer bits are needed after compression. Therefore, inspired by the rate control study \cite{Q-domain}, we interpret the reordering process of PLAS as a form of bitrate pre-allocation across different GS attributes. In light of the lack of a universal rate control approach \cite{yifan-gstrans} or loss function \cite{mped} for GS attributes, we use the initial attribute data volume and the quantization parameter (QP) as the priors to infer weighting factors for the reordering operation.  Second, the 2D map reordering process is different for I- and P-frames. For I-frames, P-MiniPLAS inherits the strategy of the two-stage Morton scan and MiniPLAS \cite{LGSCV} to generate the block-wise 2D maps (see section \ref{sec:perli_lgscv} for details), while the prior-based weighting factors are introduced to enhance the MiniPLAS performance. For P-frames, we initialize block-wise 2D maps via the two-stage Morton scan first, followed by Inter-MiniPLAS to refine them. Specifically, Inter-MiniPLAS uses the I-frame 2D maps as an anchor and predefines a maximum block size (MBS). A finite set of block sizes, selected based on MBS and conforming to the coding unit (CU) sizes supported by the standard video codecs, is used to refine P-frame block-wise maps. Using P-MiniPLAS, a small scale of reordering can generate close I- and P-frame GS maps, leading to highly efficient inter-prediction. 

For the compression module, we propose a content-aware generalized compression (CA-GC), in which two techniques are introduced. CA-GC is based on the compression framework designed by GSCV \cite{GSCV}, in which 20 BD and 10 BD are used to quantize geometry coordinates and other attributes (see section \ref{sec:perli_gscv}). GA-GC replaces the uniform 10:10 coordinate split with a non-uniform split, such as 16:4 and 12:8. The most significant bit (MSB) maps use video coding, while the least significant bit (LSB) coding choice can depend on the compression mode and split. Second, we couple the encoder's maximum CU selection with the MBS selection in P-MiniPLAS, achieving stable and robust encoding complexity control of real compression. We also optimize the implementation of CA-GC to be compatible with multiple prevalent video codecs and platforms, including HM18.0 \cite{HM}, VTM23.11 \cite{VTM}, FFMPEG libx265, and FFMPEG libx264 \cite{ffmpeg}.

Based on the proposed GSCV++, we conducted a serial experiment and ablation study on both the MPEG and Audio Video Coding Standard Workgroup of China (AVS) GS sequences. The results show that GSCV++ can report significant gains in both intra- and inter-prediction modes compared to the SOTA method; e.g., the gains over GSCV are approximately 39.6\% and 68.7\% on MPEG data in all-intra and inter modes, respectively. Besides, GSCV++ requires less time to generate GS 2D maps, decreasing from 24.78 to 4.58 seconds for I-frames and from 74.60 to 14.67 seconds for P-frames. Last but not least, GSCV++ reports faster encoding and decoding speeds than GSCV and GSCodec Studio when using the same type of downstream codecs, indicating that the 2D videos generated by GSCV++ are smoother. The contributions are summarized as follows:

\begin{itemize}
 \item We propose a comprehensive 3D GS sequence compression method, GSCV++, which can achieve SOTA compression performance, as well as lightweight 2D map generation and fast coding speed.
 \item GSCV++ applies optimizations across all stages of the GS sequence compression: EA-SHDR for preprocessing, P-MiniPLAS for 3D GS to 2D mapping, and CA-GC for the compression module. Stable and efficient compression of GS sequences is achieved even when tracked information is unavailable.
 \item We perform extensive experiments to demonstrate the effectiveness, robustness, and generalization of GSCV++. It consistently delivers strong compression performance across varying datasets and compression configurations.
\end{itemize}

The remainder of this paper is organized as follows. Section \ref{sec:related} presents the related work on 3D GS compression. Section \ref{sec:preli} introduces the background used in this paper, including the PLAS algorithm, two-stage Morton scan, MiniPLAS, and the GSCV compression pipeline. Section \ref{sec:method} details the design of GSCV++. Section \ref{sec:exp} presents the experimental results and examines the evaluated design choices. Section \ref{sec:conclusion} concludes the article.

\section{Related Work}\label{sec:related}
In this section, we first introduce GS compression in terms of both A-3DGS and I-3DGS tracks in academia, then summarize the explorations of the MPEG GSC working groups.

\subsection{A-3DGS: Optimization-Based GS Compression}
In the academic community, A-3DGS methods are commonly referred to as optimization-based or generative compression methods. The generation process of 3D GS offers considerable scope for more compact data representations, since the mapping from 3D GS representations to rendered images is not one-to-one: distinct 3D GS representations may yield similar rendered content and quality. Therefore, the input of A-3DGS methods is multiview images/videos, and a more compact GS representation is generated via an end-to-end training process. The compact representation can use the vanilla GS format or other explicit, implicit, or hybrid representations. Scaffold-GS \cite{lu2024scaffold} is an early example of this approach, in which a group of anchors was initialized with latent features. The explicit GS primitives can be reconstructed from the anchor information via MLPs. Inspired by Scaffold-GS, HAC \cite{chen2024hac} and HAC++ \cite{chen2025hac++} were proposed by introducing an effective context model based on multiresolution hash coding, which can achieve approximately 60x to 100x reduced storage. CompGS and CompGS++ \cite{liu2024compgs} also constructed compact latent features based on anchors, and both intra- and inter-prediction were implemented via anchor and residual features.  In contrast, LightGaussian \cite{fan2023lightgaussian} approached the problem from the perspective of controlling the number of primitives and proposed a rendering-based pruning method. The experiment shows that removing 50\% to 60\% of primitives can achieve similar rendering quality after a limited number of iterations of fine-tuning. EAGLES \cite{girish2025eagles} and HybridGS \cite{yanghybridgs} used low-dimensional latent features to reconstruct high-dimensional explicit GS attributes, provided some GS features are highly redundant. Besides, HybridGS realizes fast encoding and decoding by combining standard point cloud encoders with compact GS data. Codebooks and vector quantization constitute another commonly used approach, such as \cite{navaneet2025compgs, niedermayr2024compressed}. More recently, A-3DGS methods have been extended to the 4D domain to generate compact dynamic GS representations, such as 4DGC \cite{hu20254dgc}, ADC-GS\cite{huang2025adc}, and Light4GS\cite{liu2025light4gs}.

Optimization-based and dynamic GS methods can jointly redesign the representation and reconstruction processes, which can offer compression opportunities beyond coding an existing GS sequence. Generally, they have a higher requirements for encoding time, computational resources, as well as customized design. Our focus is narrower and inspired by industrial requirements: reuse standard video codecs for already constructed GS sequences without requiring scene-specific reconstruction optimization during compression. 

\subsection{I-3DGS: Optimization-Free GS Compression}
To develop more practical GS compression methods, I-3DGS methods have attracted more attention in recent times. Different from A-3DGS, which involves a time-consuming optimization process, I-3DGS aims to compress GS while retaining the same representation type: both the input and output are the generated GS data. Inspired by point cloud compression, GGSC \cite{yang2024benchmark} designed a two-branch compression pipeline: using GPCC to compress the GS coordinates and graph signal processing \cite{graphsim} related techniques to compress other attributes. HGSC \cite{huang2024hierarchical} proposed a hierarchical strategy for progressive primitive compression. FCGS \cite{chenfast} is the first method that realizes learning-based I-3DGS compression via both intra- and inter-primitive context models. 

To facilitate the deployment of the 3D GS compression method, MPEG prioritizes the lightweight I-3DGS method based on the requirement proposal from the industry \cite{m72430}: using standard video or point cloud codecs to achieve effective GS compression. GSCodec Studio \cite{li2025gscodec} was selected as the video-based GS compression anchor, while GPCC v1 \cite{GPCCv1} was regarded as the point cloud-based anchor. Specifically, GSCodec Studio uses the PLAS algorithm \cite{PLAS-code} to convert 3D GS data into smooth 2D maps, followed by an FFMPEG libx265 codec to compress these maps. The proposal \cite{wg7-73294} merges multiple PLAS maps from a single GS frame into a single video and compresses it using HM18.0 \cite{HM}. LGSCV \cite{LGSCV} combines MiniPLAS with PCA \cite{pca-wg4} and HM18.0 for efficient all-intra GS coding; it does not introduce inter-frame correspondence modeling. GSCV \cite{GSCV} is the pioneering work that studies GS sequence compression without tracked information: its Inter-PLAS matches later-frame attribute maps to an anchor when reliable primitive correspondence is unavailable. However, Inter-PLAS has very high complexity, and the performance of GSCV can still be significantly improved due to the lack of preprocessing optimization.

Although current video-based GS compression methods have demonstrated impressive compression performance, there is still room for more effective 2D map generation for both intra- and inter-prediction, as well as a more appropriate video codec configuration. This motivates the design of GSCV++ in this paper.

\section{Preliminary}\label{sec:preli}
In this section, we review three relevant components underlying GSCV++. First, we introduce the implementation of PLAS, which is the basis of the GS 3D-to-2D map generation. Second, the two-stage Morton scan and MiniPLAS are summarized, which form the basis of the proposed P-MiniPLAS.  Third, the compression module designed in GSCV is used as the basic framework of GSCV++, leading to a more detailed description in this section.
\subsection{Brief Introduction of PLAS}\label{sec:preli_plas}
PLAS \cite{PLAS} is a heuristic algorithm that reorders GS primitives into smooth 2D attribute maps. The feature matrix $G\in\mathbb{R}^{n\times59}$ contains $n$ primitives with 3 coordinate, 3 SH DC, 45 SH AC, 1 opacity, 3 scale, and 4 rotation channels. Minimal pruning based on scale and opacity fits the primitives to a square grid, followed by random initialization of $I_0=I\in\mathbb{R}^{M\times M\times m}$, where $m\leq59$ channels guide smoothness evaluation. A common permutation preserves alignment across all attribute channels without additional signaling.

PLAS uses progressively smaller Gaussian-blur radii $r_i=\lfloor r_0\tau^{i-1}\rfloor$, where $r_0=M/2-1$, $\tau\in(0,1)$ is the decay factor, and the minimum radius is 1. At level $i$, the blur-kernel width is $K_i=\min(M-1,2r_i+1)$ and the permutation-block width is $B_i=\max\{B_{min},2\lfloor\min(K_i+1,M)/2\rfloor\}$, with $B_{min}=16$ in vanilla PLAS. The target map is the Gaussian-blurred input, $T_i=blur(I_{i-1},K_i)$.

The input and target maps are partitioned into corresponding $B_i\times B_i$ blocks for batched processing. Within each complete block, whose even width allows grouping into sets of four pixels, PLAS evaluates all $4!=24$ permutations of each randomly selected group and retains the assignment with the smallest L2 loss to the corresponding target pixels. The resulting map $I_i$ is passed to the next level. Non-increasing $K_i$ and $B_i$ yield coarse-to-fine refinement toward smooth attribute maps.

\subsection{Two-Stage Morton Scan and MiniPLAS}\label{sec:perli_lgscv}
\subsubsection{Two-Stage Morton Scan}
Given spatial coordinates $(x,y,z)$, a 3D Morton code interleaves the binary representations of the coordinates, which can permute the spatially adjacent primitives into close indices \cite{Morton}. Attributes of 3D GS primitives exhibit local spatial similarity, indicating that after 3D Morton sorting, primitives with adjacent indices tend to share similar attribute values. Then, the scan order of the 2D Morton code is used to generate 2D maps. The generated 2D maps exhibit noticeable block structures, with smooth content within each block, making them well-suited for video codecs.
\subsubsection{MiniPLAS}
MiniPLAS improves vanilla PLAS by defining a group of new block sizes for pixel permutation: the block sizes of permutation are defined as $4n\times4n$, $n=1,2,4, ...$. The selected block sizes are the CU or sub-CU sizes in video codecs, such as $8\times8$. MiniPLAS is also structured in a progressive manner. Based on the ablation study in LGSCV \cite{LGSCV}, for single-frame compression in all-intra mode, using a block size of $8\times8$ for two iterations is good enough, which is also adopted in the proposed GSCV++ I-frame 2D map generation. 

\subsection{Compression Module of GSCV}\label{sec:perli_gscv}
GSCV \cite{GSCV} combines 8 frames into one GoP, with the first frame designated as the I-frame and the remaining 7 frames as non-anchor frames (called P-frames here for map generation, irrespective of their codec picture types). 20 BD is used to quantize the coordinates and 10 BD is applied to quantize other attributes. For each GS frame, GSCV generates 1, 15, and 1 frame images for the color SH DC, SH AC, and scale. For opacity and rotation, GSCV uses the mid-value of BD to pad two channels to form 1 and 2 images, respectively. Most video codecs cannot compress 20 BD images directly. GSCV splits the coordinate map into 2 images, namely the high part and the low part of the data. Each part takes 10 BD. For each GS frame, 22 images are generated. GSCV extracts the corresponding images from each frame within a GoP to form 22 videos and saves them in YUV444 format.  Lossless compression is used for coordinates, while other attributes employ the lossy mode. Note that GSCV directly considers the video data as YUV rather than RGB, omitting the operation that converts data from RGB to the YUV color space for YUV file generation. More details can be found in \cite{GSCV}.

\section{Method}\label{sec:method}
\begin{figure*}[t] {
\centering
    \setlength{\abovedisplayskip}{0pt}
    \setlength{\belowdisplayskip}{0pt}
    \includegraphics[width=1\textwidth]{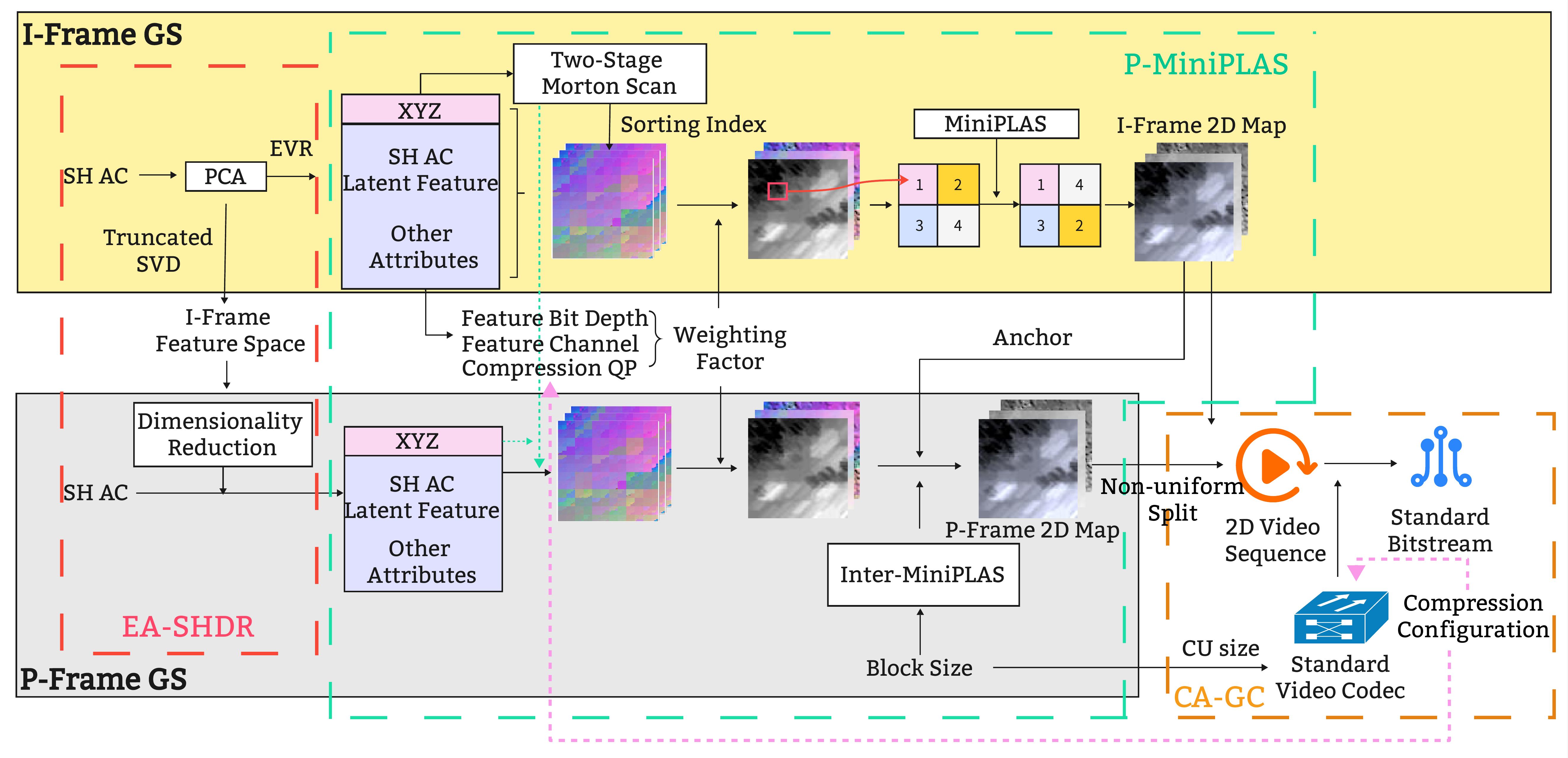}
    \caption{Framework of GSCV++.}
    \label{fig:framework} }
\end{figure*}

 Fig. \ref{fig:framework} shows the framework of GSCV++, in which the main contributions are in three modules: EA-SHDR, P-MiniPLAS, and CA-GC. EA-SHDR belongs to the preprocessing step illustrated in Fig. \ref{fig:motivation}, where we propose a lightweight method based on PCA to achieve color SH AC dimensionality reduction for inter-frame mode. P-MiniPLAS corresponds to the 3D-to-2D mapping procedure. Two technologies are included: 1) a prior-based PLAS weighting factor inference algorithm designed to assign different smoothness levels for different GS attributes, leading to bitrate pre-allocation; 2) a weighted version of MiniPLAS used for I-frame 2D map generation and an inter-MiniPLAS proposed for P-frame 2D map generation. CA-GC is an optimized compression module, in which we design a non-uniform BD split and a differentiated compression scheme for the GS geometric coordinate. The block sizes used in P-MiniPLAS are specified in the video codec CU configuration, enabling the development of a complexity-controlled compression module with a fast encoding speed. We will detail the proposed GSCV++ in the following part of this section.

\subsection{Energy-Adaptive Inter-Frame SH AC Dimensionality Reduction (EA-SHDR)}
Previous studies \cite{LGSCV, pca-wg4} have shown that the GS SH AC component is highly redundant; introducing dimensionality reduction before compression can yield significant RD gains. To facilitate GS sequence compression and improve the inter-prediction of SH AC, GSCV++ adopts different strategies for I- and P-frames. For an I-frame with $N$ GS primitives, let $G^{I}_{AC}\in\mathbb{R}^{N\times45}$ contain its 45 SH AC feature channels. Unlike LGSCV, which fixes the latent dimension, GSCV++ selects it using the cumulative explained variance ratio (EVR) of the I-frame features. The reduction process consists of three steps:

\begin{subequations}\label{eq:pca}
\textit{i) Centering and basis estimation.} First, compute the per-channel mean $\boldsymbol{\mu}^{I}=mean(G^{I}_{AC})\in\mathbb{R}^{1\times45}$. Subtract this mean from every primitive and apply SVD:
\begin{align}
    \overline{G^{I}_{AC}}&=G^{I}_{AC}-\mathbf{1}_N\boldsymbol{\mu}^{I}, \quad
    \overline{G^{I}_{AC}}&=\mathbf{U}^{I}\boldsymbol{\Sigma}^{I}(\mathbf{V}^{I})^{\mathsf{T}}.
    \label{eq:pca_basis}
\end{align}
Here, $\mathbf{1}_N$ is an $N$-dimensional all-ones column vector, and $\mathbf{U}^{I}$ and $\mathbf{V}^{I}$ contain the left and right singular vectors. The diagonal entries $\sigma_i^{I}$ of $\boldsymbol{\Sigma}^{I}$ are the singular values in descending order. The corresponding columns $\mathbf{v}_i\in\mathbb{R}^{45}$ of $\mathbf{V}^{I}$ are the principal directions.

\begin{samepage}
\textit{ii) Energy-adaptive dimension selection.} The variance along principal direction $\mathbf{v}_i$ is $\lambda_i=(\sigma_i^{I})^2/(N-1)$. Given a target retained-variance fraction $A\in(0,1]$, select the fewest leading components whose cumulative explained variance reaches $A$:
\begin{equation}
    k^\ast=\min\left\{d\in\{1,\ldots,45\}\;\middle|\;\frac{\sum_{i=1}^{d}\lambda_i}{\sum_{i=1}^{45}\lambda_i}\ge A\right\}.
    \label{eq:pca_dimension}
\end{equation}
Thus, $k^\ast$ adapts to the SH AC variance distribution of the I-frame.
\end{samepage}

\textit{iii) Three-channel packing and projection.} Round the retained dimension up to $k=3\lceil k^\ast/3\rceil$ so that the latent channels form three-channel images without padding. Using the first $k$ principal directions, define $\mathbf{V}_{k}^{I}=[\mathbf{v}_1,\ldots,\mathbf{v}_k]\in\mathbb{R}^{45\times k}$ and project the centered features:
\begin{equation}
    z^{I}_{AC}=\overline{G^{I}_{AC}}\mathbf{V}_{k}^{I}\in\mathbb{R}^{N\times k},
    \label{eq:pca_projection}
\end{equation}
where $z^{I}_{AC}$ is the reduced I-frame SH AC feature matrix.
\end{subequations}

Applying PCA independently to P-frames can produce different principal directions, hindering attribute matching in P-MiniPLAS and the subsequent inter-prediction process. We therefore reuse both the I-frame mean $\boldsymbol{\mu}^{I}$ and basis $\mathbf{V}_{k}^{I}$ to express the P-frame SH AC features in the same coordinate system:
\begin{equation}
    z^{P}_{AC}=(G^{P}_{AC}-\mathbf{1}_N\boldsymbol{\mu}^{I})\mathbf{V}_{k}^{I}\in\mathbb{R}^{N\times k}.
    \label{eq:pca_pframe}
\end{equation}
The decoder reconstructs SH AC features as $\widehat{G}^{t}_{AC}=\widehat{z}^{t}_{AC}(\mathbf{V}_{k}^{I})^{\mathsf{T}}+\mathbf{1}_N\boldsymbol{\mu}^{I}$ for both I- and P-frames. This method facilitates GS inter-frame compression and reduces preprocessing time by reusing the I-frame PCA results. The decoder uses the same basis and mean for all frames in the GoP. 

\subsection{Prior-Based Inter-MiniPLAS Refinement (P-MiniPLAS)}
3D GS to 2D mapping is a key step because the characteristics of the generated 2D video will directly influence the effectiveness of the downstream standard video codecs. The proposed P-MiniPLAS module is an enhancement of PLAS \cite{PLAS-code} that adds attribute-aware weighting factors and inter-frame refinement. 

\subsubsection{Prior-based PLAS Weighting Factor Inference}\label{sec:plasWeight}
For the I-frame GS samples, we first use the two-stage Morton scan \cite{LGSCV} to generate block-wise 2D images and scale them into the same value range, e.g., [0,255]. Unlike LGSCV, which directly performs MiniPLAS on these 2D images, a weighting factor is calculated for each GS attribute to rescale its image values. These weighting factors act as feature-scaling coefficients that can guide the subsequent permutation process, prioritizing the more important attributes instead of uniformly handling all attributes in terms of texture smoothness. There are two types of prior knowledge used to infer these weighting factors: initial attribute data volume and target bitrate. 

First, assuming the same degree of lossy compression, e.g., the same QP, the larger initial data volume generally corresponds to a larger proportion of the final bitstream. If the permutation process of PLAS can generate smoother images for the attributes that originally contain larger amounts of data, a substantial reduction in the overall bitrate can be achieved.  Second,  for 3D GS, different attributes vary in their importance for rendering quality, as well as in their sensitivity to lossy compression. There is a consensus that geometric coordinates play the most critical role, and scale and opacity are more sensitive to lossy compression than color SH. However, we still lack a generalized model for quantifying the relative importance of different GS attributes. Nevertheless, valuable insights can be drawn from the QP design in MPEG GSC research: attributes of higher importance are typically assigned smaller QP values to achieve a more favorable RD trade-off.

Therefore, the weighting factors can be formulated as follows:
\begin{align}
    W_a = f^{1}(D^a)\cdot f^{2}(QP^a),
\end{align}
where $a$ represents the specific attributes, such as SH AC and opacity, $D^a$ and $QP^a$ are the initial data volume and QP of the corresponding attribute. $f^1(\cdot)$ and $f^2(\cdot)$ are the mapping functions to derive the quantified parameters. For $f^1(\cdot)$, we define 
\begin{align}
    f^{1}(D^a) = BD^a \cdot CH^a,
\end{align}
where $BD^a$ and $CH^a$ are the BD and channel numbers of attribute $a$. 

The calculation of $f^2(\cdot)$ is inspired by Q-domain rate-distortion analysis from video coding \cite{Q-domain}. The rate $R$ generally decreases as the QP increases. However, research shows that the relation between $R$ and $1/Q_{step}$ can be more accurately approximated by a linear model rather than the relation between $R$ and $1/QP$, therefore, we set
\begin{align}
    &f^{2}(QP^a) = \frac{1}{Q_{step}^a} = 2^{\frac{4-QP^a}{6}}, \\
    &R^a \approx K^a\,f^{2}(QP^a)^{\beta^a} = K^a\,2^{\beta^a(4-QP^a)/6},
\end{align}
where $R^a$ is the estimated number of coded bits for attribute $a$. $f^{2}(QP^a)$ shows a positive correlation with $R^a$ but is not strictly proportional, as $R^a$ is additionally influenced by content-dependent hyperparameters $K^a$ and $\beta^a$. However, for our work, this correlation provides a sufficient, computationally inexpensive basis for the weighting factor; therefore, we simply assume $f^{2}(QP^a) \propto R^a$. Precisely computing $R^a$ incurs a significant computational overhead, as PLAS is a multi-iteration optimization process.

As a bitrate-allocation prior, $W_a$ scales the normalized feature values before evaluating squared Euclidean distances, its contribution to the permutation cost is proportional to $W_a^2$. After we obtain $W_a$ for each GS attribute, we use MiniPLAS to refine I-frame block-wise images to obtain smooth 2D maps. Given different QPs, we will generate different 2D maps before compression. If the encoding configuration is not available before 2D map generation, using only the initial data volume also yields a clear gain over equal weighting factors, as shown in Section \ref{sec:ab_plasweight}. It indicates that the proposed method is flexible and scalable.

\subsubsection{Inter-MiniPLAS for P-frame Refinement} \label{sec:inter-miniplas}  \begin{figure*}[!t] \centering \includegraphics[width=\textwidth]{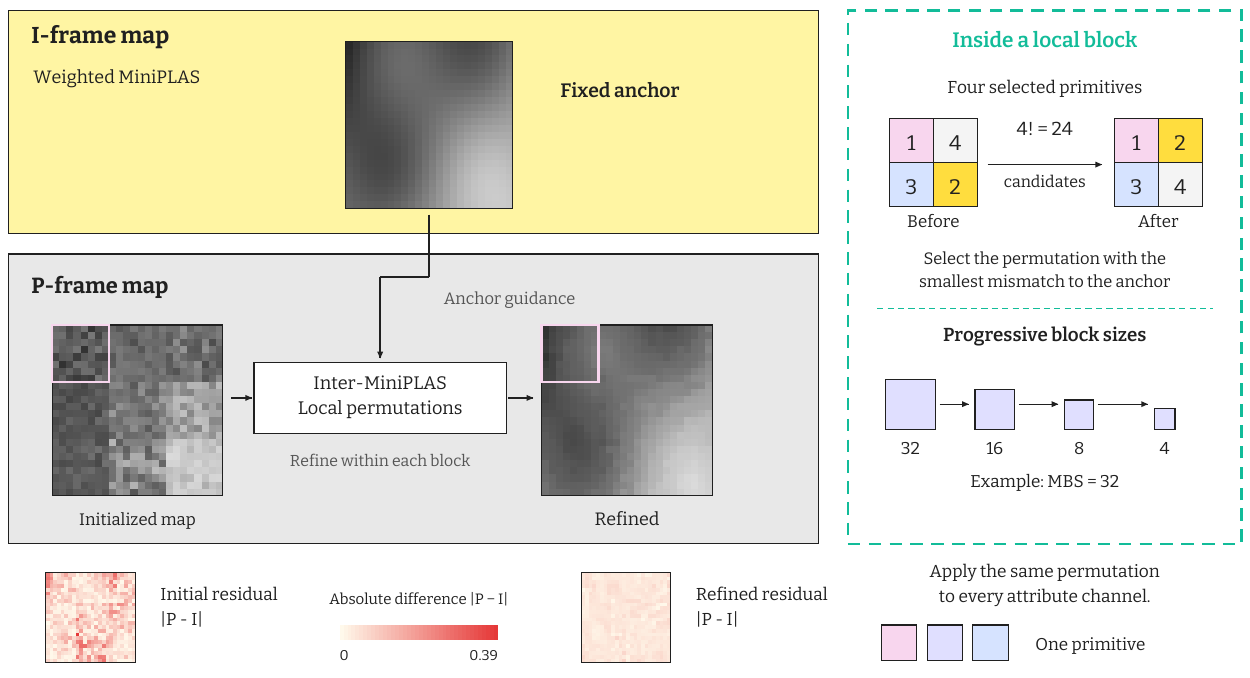} \caption{Illustration of Inter-MiniPLAS. The I-frame map is held fixed while the initialized P-frame map is refined through local permutations of four primitives and progressively smaller blocks. The same permutation is applied to all attribute channels.} \label{fig:inter_miniplas_schematic} \end{figure*}

Theoretically, given I-frame maps as anchors, directly using the Inter-PLAS \cite{GSCV} can generate P-frame maps similar to the I-frame maps if a sufficient number of iterations are conducted. Using four selected positions $p_1,\ldots,p_4$ within a block as an example, let $\widetilde{\mathbf{g}}^I$ and $\widetilde{\mathbf{g}}^P$ denote the normalized, weighted feature vectors. The selected assignment is $\pi^\ast=\arg\min_{\pi\in S_4}\sum_{j=1}^{4}\|\widetilde{\mathbf{g}}^P(p_{\pi(j)})-\widetilde{\mathbf{g}}^I(p_j)\|_2^2$, where $S_4$ contains all 24 permutations. Considering a 2D map derived from a GS with more than one million primitives, carrying out a comprehensive exhaustive search of all permutations is extremely costly, which can explain why the Inter-PLAS algorithm has a substantially higher computational cost.

Therefore, for P-frame GS samples, we propose an Inter-MiniPLAS algorithm to generate P-frame 2D maps, as Fig. \ref{fig:inter_miniplas_schematic} illustrates. After using a two-stage Morton scan to initialize P-frame 2D maps, we propose progressive local permutations to remap them with I-frame MiniPLAS maps as anchors. The progressive block sizes are small, such as 32x32 and 4x4. The reason is that the two-stage Morton scan already uses the spatial distribution to initialize block-wise 2D maps \cite{LGSCV}, it is unnecessary to permute the pixels from a very large receptive field. We consequently restrict the permutation search to a small local region, which can accelerate convergence without disrupting the local block correlation. Besides, the downstream video codecs use square pixel blocks of certain sizes as coding units (CU). In the absence of explicit motion information, favorable compression performance can be achieved as long as the pixel blocks at corresponding positions are sufficiently similar. 

Standard video codecs use a coding tree unit (CTU) as the largest basic processing unit, such as $64\times64$ in HEVC and $128\times128$ in VVC. We define the maximum permutation block size (MBS) as $B_{MBS}\times B_{MBS}$, $B_{MBS}\in \{\rm CTU\ Size\}$. Considering that the PLAS algorithm is a progressive process,  the block size used in the $i$-th iteration for the P-frame is selected as 
\begin{align}
    B_i = \frac{B_{MBS}}{2^i}, i = 0, 1, ..., 
\end{align}
where $B_i$ decreases to 4 as the minimum block size. For example, when $B_{MBS} = 32$, there are four block-size levels with $B_i = 32,\ 16,\ 8,\ 4$. Each level can contain multiple permutation sweeps. 

For an $M\times M$ map, the permutation-enumeration cost of one sweep at block size $B_i$ is $O(\lfloor \frac{M}{B_i} \rfloor^2\times(\frac{B_i^2}{4}\times24))$ for a single feature channel, where $\lfloor \cdot \rfloor$ is the floor function. $\lfloor \frac{M}{B_i} \rfloor^2$ is the number of blocks in the 2D map, and $\frac{B_i^2}{4}\times24$ is the number of permutations within each block, because PLAS uses 4 pixels as one permutation group (4! = 24 permutations for each unit). Therefore, assuming one complete sweep per block-size level, the overall computation of P-frame Inter-MiniPLAS for one channel of data is $O(\Sigma_{i=0}^{l}\lfloor \frac{M}{B_i} \rfloor^2\times(\frac{B_i^2}{4}\times24))$, $l = \log_2 B_{MBS} -2$. When $B_{MBS}\ll M$, the computational complexity can be simplified and estimated by $O(M^2\log B_{MBS})$. If level $i$ requires $s_i$ sweeps over $C$ feature channels, the permutation cost becomes $O(CM^2\sum_{i=0}^{l}s_i)$. The measured runtimes are reported separately in Table \ref{tab:time}.

The block size setting of Inter-MiniPLAS not only reduces the computation time and complexity for the 2D map generation, but it can also be used as prior information for downstream video codec configuration. We will detail this part in the next section.

\subsection{Content-Aware Generalized Compression (CA-GC)}
We use the compression framework proposed in GSCV as the basis of CA-GC, and introduce two new techniques for efficient GS map compression. The first technique focuses on geometric coordinates, because coordinate data have a relatively high BD and are encoded in a lossless manner, which causes them to occupy a substantial portion of the total bitrate. In particular, in low-bitrate regimes, the bitrate contribution of the coordinate can exceed half of the overall bitrate, as discussed in section \ref{sec:exp_rate-allo}. Therefore, targeted optimization for coordinate compression can yield significant benefits. The second technique builds the connection between the MBS of Inter-MiniPLAS and the maximum CU size of the codec, which can significantly reduce the coding time without sacrificing obvious coding gains.

\subsubsection{Differentiated Encoding Strategy for Geometric Coordinates} \label{sec:diff_xyz}

\begin{figure}[t] {
\centering
    \setlength{\abovedisplayskip}{0pt}
    \setlength{\belowdisplayskip}{0pt}
    \includegraphics[width=\columnwidth]{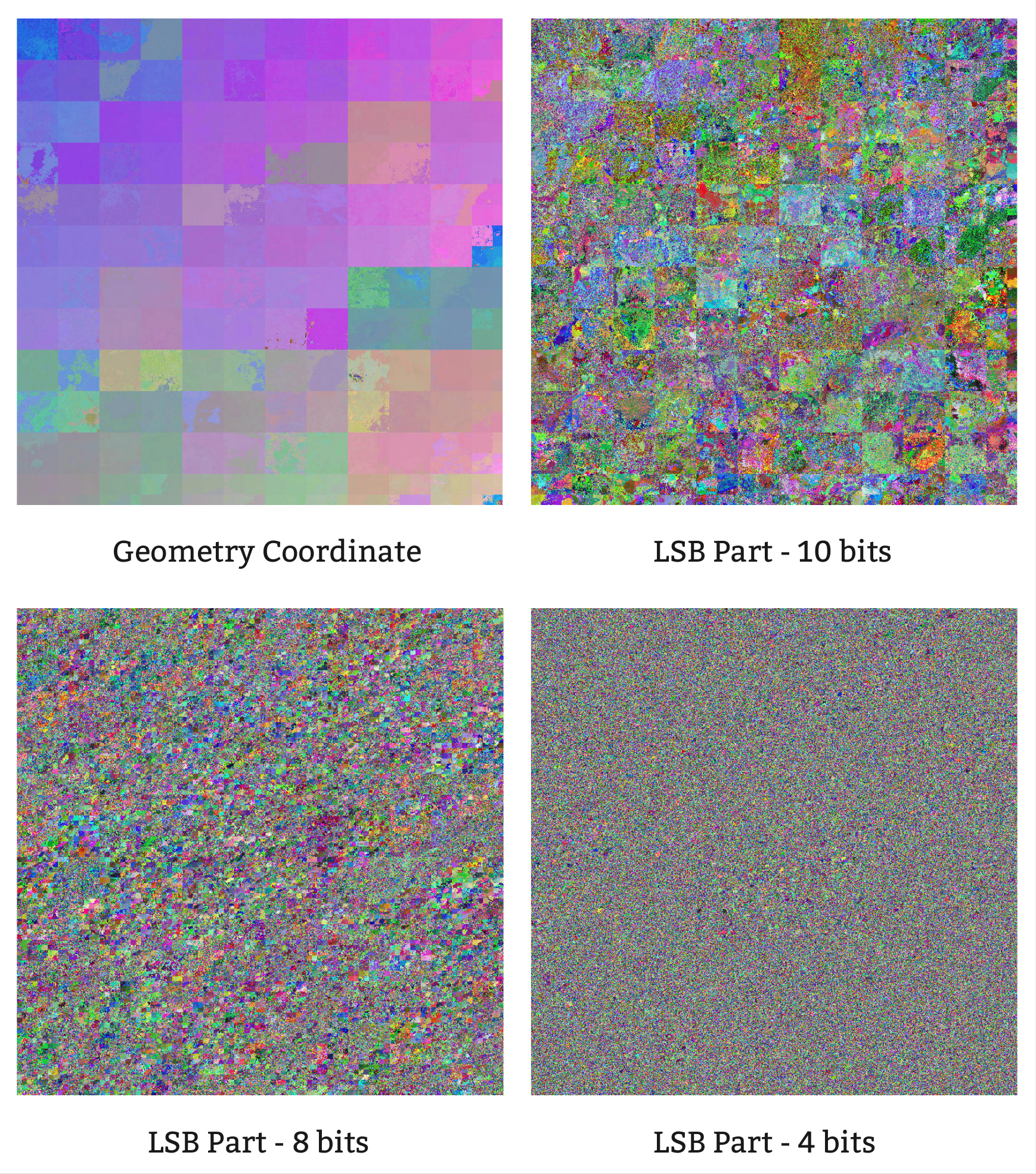}
    \caption{Snapshot of the geometric coordinate and its LSB part.}
    \label{fig:xyz_show} }
\end{figure}

Although GSCV splits the 20-bit coordinate map into two 10 BD maps, which are compatible with standard video codecs, this operation fails to deliver effective compression results. The reason is that the LSB part is highly irregular, as shown in Fig. \ref{fig:xyz_show} upper right, leading to limited performance when using video codecs. However, for 10 BD LSB, we can still find some contours that satisfy the block-wise style, indicating that 10 BD LSB images exhibit some low-frequency information.  

We use a non-uniform coordinate split to assign more bits to the MSB maps and leave fewer bits in the LSB maps. The bottom panels of Fig. \ref{fig:xyz_show} illustrate the less structured appearance of the 8-bit and 4-bit LSB maps. Experimental results show that standard video codecs can report better lossless compression performance for non-uniform coordinate maps than for uniform splits. Besides, for noise-like LSB maps, video codecs might not be necessary, and we can consider some more lightweight compression algorithms, such as LZMA \cite{ziv2003universal}. Sometimes, we do not even need to conduct real compression for LSB due to very limited compression room. Section \ref{sec:diff_xyz_ablation} compares LZMA with lossless HM18.0 and gives mode- and split-dependent recommendations.

\subsubsection{Encoding Complexity Control via CU Size}
In standard video codecs, given a predefined maximum CU size, the RD optimization (RDO) algorithm is conducted to determine the best CU partition and compression modes, which is an extremely time-consuming process due to the need to assess all strategies. The local refinement of Inter-MiniPLAS motivates using MBS as a prior for limiting the maximum CU size. Given $B_{MBS}$, we select the closest supported maximum CU size for compression. Note that using larger $B_{MBS}$ and CU size cannot always provide better RD results than the smaller ones: excessively large $B_{MBS}$ can severely disrupt the smooth geometric coordinate maps generated by the two-stage Morton scan, as well as incur poorer overall bitrate distribution. In the experiments, we evaluate this design choice and quantify the reduction in encoding complexity.

\section{Experiments}\label{sec:exp}
In this section, we report the performance of the proposed GSCV++ on both datasets provided or derived by MPEG and AVS. Besides, we conduct a comprehensive ablation study to demonstrate the effectiveness of different modules.
\subsection{Experimental Settings}
\subsubsection{Dataset Selection and Construction}

For the MPEG datasets, we select three GS sequences, including ``bartender'', ``breakfast'', and ``cinema'', with tracked and semi-tracked versions from the MPEG GSC working groups. Each sequence has 32 frames; the main experiments use the first eight frames as one GoP to test RD performance (Larger GoP sizes are reported in Section \ref{sec:ab_gop}). ``Tracked'' means all the GS frames are refined from a shared canonical GS model, which was trained using the merged point cloud with ground truth images from all viewpoints and time instants. Densification is disabled during the refinement process, ensuring that all the frames have the same number of primitives. ``semi-tracked'' means densification is allowed during the finetuning process, which leads to a different number of primitives for different frames. 
\begin{figure}[!t] {
\centering
    \setlength{\abovedisplayskip}{0pt}
    \setlength{\belowdisplayskip}{0pt}
    \includegraphics[width=\columnwidth]{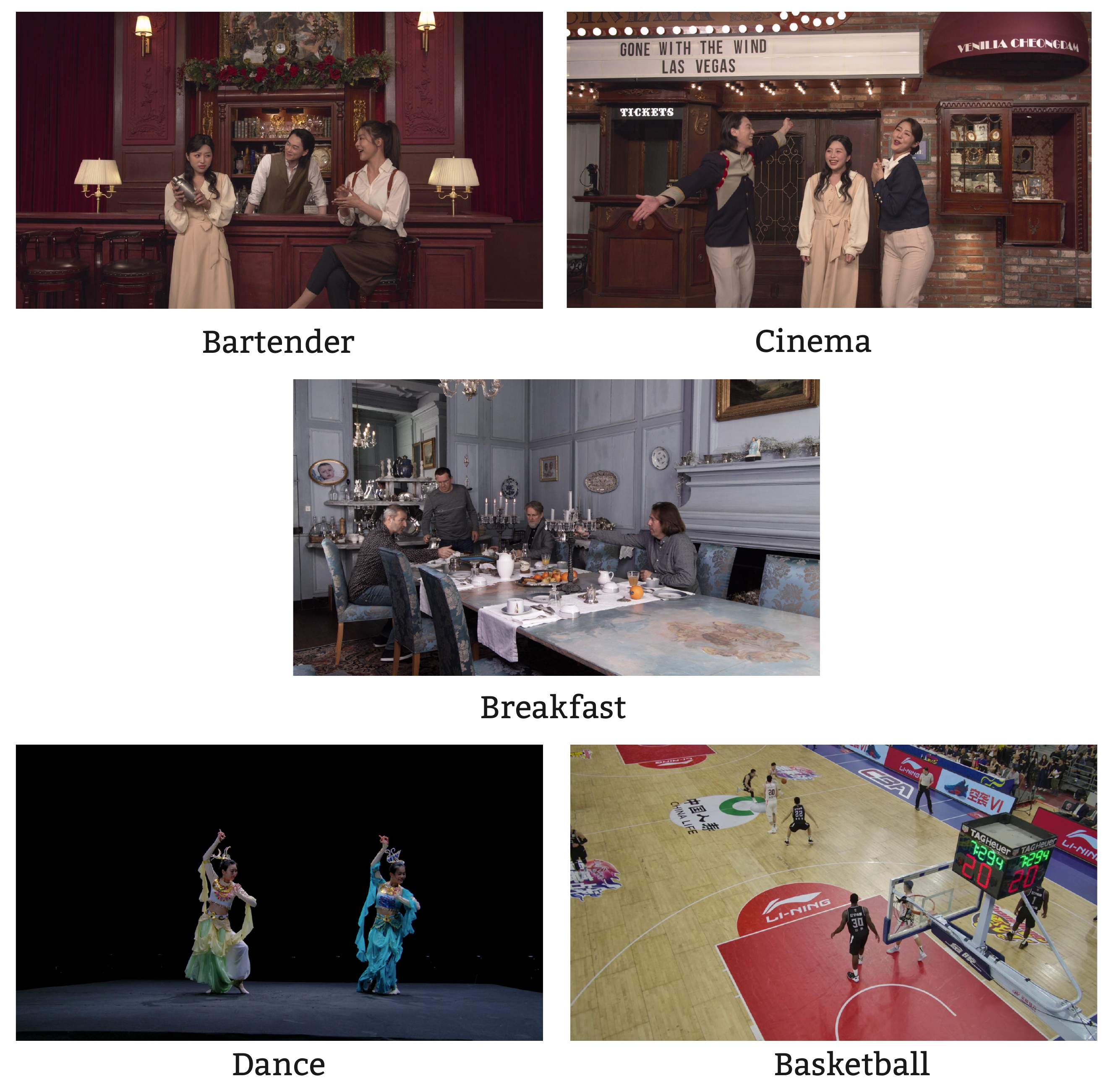}
    \caption{Snapshots of the datasets.}
    \label{fig:dataset} }
\end{figure}
For AVS datasets, we select two dynamic scenes that exhibit clear differences from those provided by MPEG, i.e., ``dance\_dunhuang\_pair''(dance) and ``basketball'' from AVS-VRU \cite{dg4}. We follow the instructions of the MPEG document \cite{dataset} to generate GS sequences. Snapshots of the datasets are shown in Fig. \ref{fig:dataset}.

\begin{figure*}[t]
\centering
\begin{minipage}{0.24\textwidth}
    \includegraphics[width=\linewidth]{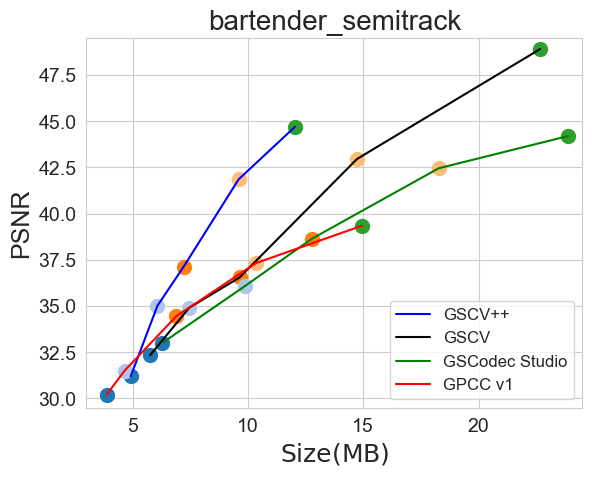}
\end{minipage}
\begin{minipage}{0.24\textwidth}
    \includegraphics[width=\linewidth]{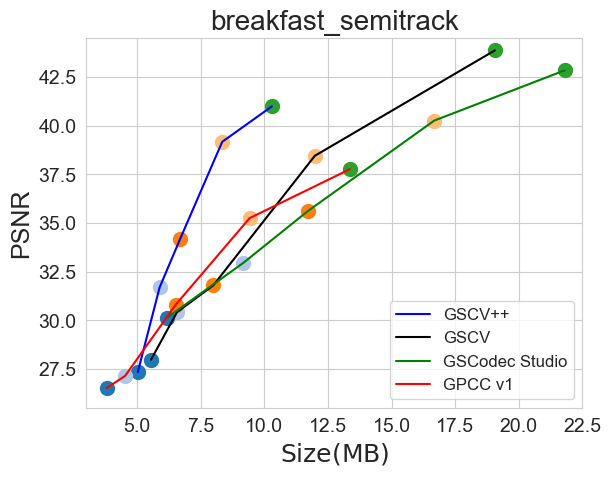}
\end{minipage}
\begin{minipage}{0.24\textwidth}
    \includegraphics[width=\linewidth]{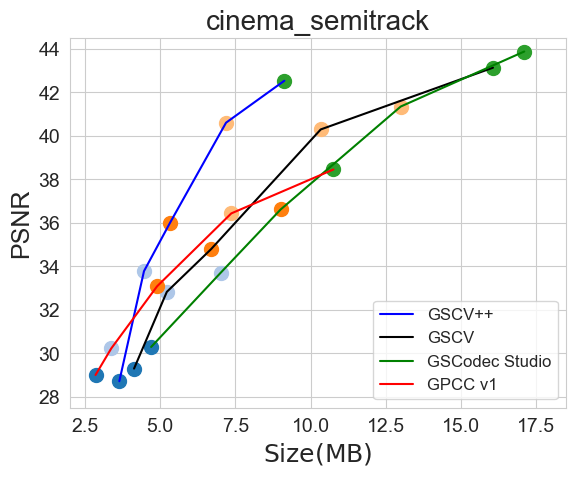}
\end{minipage}
\begin{minipage}{0.24\textwidth}
    \includegraphics[width=\linewidth]{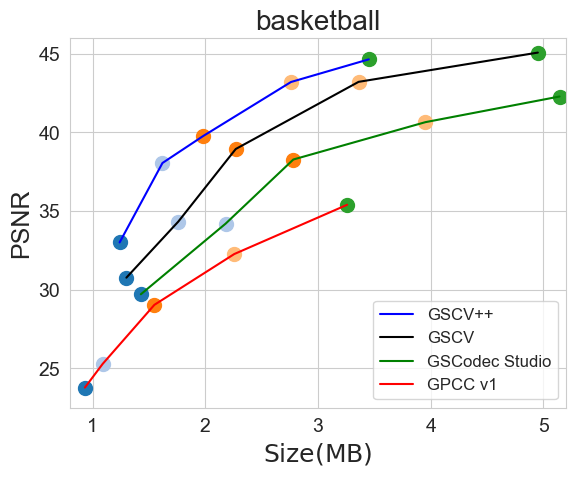}
\end{minipage}\\
\begin{minipage}{0.24\textwidth}
    \includegraphics[width=\linewidth]{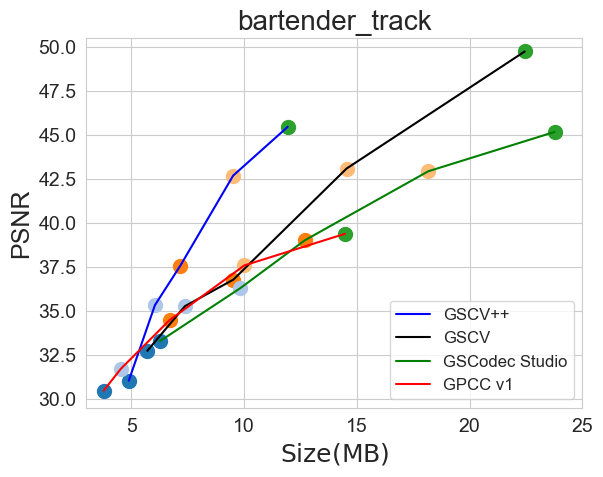}
\end{minipage}
\begin{minipage}{0.24\textwidth}
    \includegraphics[width=\linewidth]{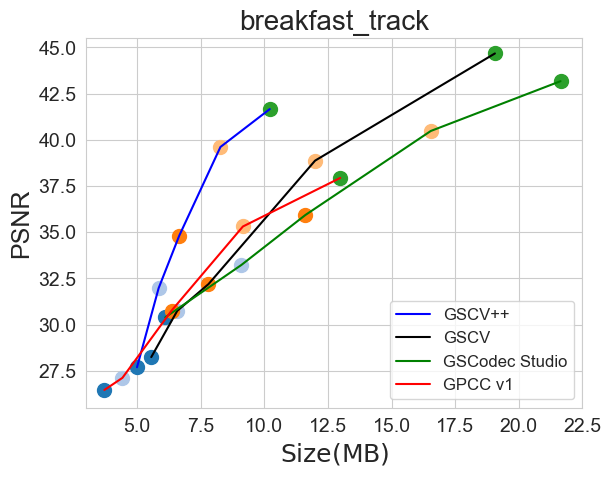}
\end{minipage}
\begin{minipage}{0.24\textwidth}
    \includegraphics[width=\linewidth]{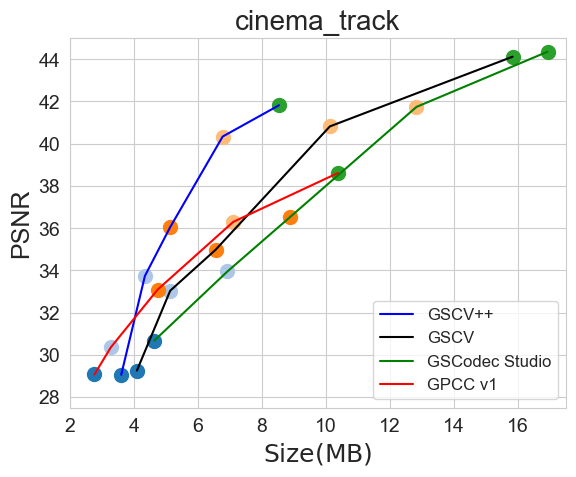}
\end{minipage}
\begin{minipage}{0.24\textwidth}
    \includegraphics[width=\linewidth]{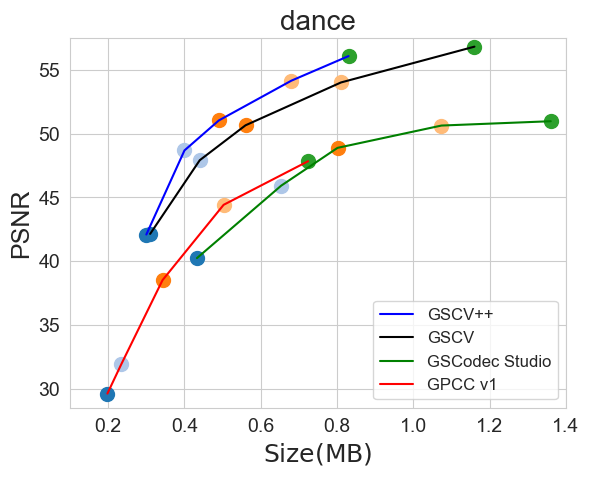}
\end{minipage}
\caption{Intra-prediction performance comparison on the MPEG and AVS dataset.}
\label{fig:mian_intra}
\end{figure*}

The target condition is unavailable inter-frame primitive tracked information. To construct this condition from tracked sequences, we shuffle each frame independently before compression; the label tracked below describes the source data, not the preserved input indices. For the semi-tracked sequences, LightGaussian pruning \cite{fan2023lightgaussian} produces the equal primitive counts required by the evaluated fixed-resolution map pipeline. 

\subsubsection{Comparison Method}

We compare the proposed GSCV++ against two anchors adopted in the MPEG GSC working group, GSCodec Studio \cite{li2025gscodec} and GPCC v1 \cite{GPCCv1}, as well as the SOTA video-based GS sequence compression method GSCV, which also addresses the problem of unavailable inter-frame correspondence. GPCC v1 does not support GS inter-prediction, so we use all-intra mode for all the testing conditions. GSCodec Studio \cite{li2025gscodec} pipeline does not support the no-tracked cases, which is also one of the motivations for this paper. We therefore exclude GSCodec Studio and GPCC v1 from RD gains and do not use their results to rank compression performance. Following \cite{N0676}, quality is evaluated at views 9 and 11 for ``bartender'' and ``cinema'', and views 6 and 8 for ``breakfast''. The AVS sequences use views 1, 11, 21, 31, 41, and 51 for ``dance'', and views 1, 11, 21, and 31 for ``basketball''.

\subsubsection{Model Parameters}\label{sec:exp_mp}

For the EA-SHDR module, we use an energy threshold $A=0.99$ to implement color SH AC dimensionality reduction. To avoid padding the latent feature $z_{AC}$ when generating three-channel images, we use $k=3\lceil k^\ast/3\rceil$. For example, $k^\ast=20$ in Eq. \eqref{eq:pca} gives $k=21$. For the P-MiniPLAS module, when calculating the PLAS weight $W_a$ for coordinate lossless compression, we set $QP=QP_{DC}$ to calculate $f^2(QP^a)$, then scale this value up by a factor of 1.25. The reason is that color SH DC usually uses a small QP.  We expect a relatively larger weighting factor for the coordinate, considering that lossless compression generally leads to a larger bitstream. Other coordinate-weight settings are discussed in Section \ref{sec:ab_plasweight}. We set the block size of I-frame MiniPLAS to [8,8] as suggested by \cite{LGSCV}, while the MBS for the P-frame Inter-MiniPLAS is set to 64. For the CA-GC module, as with GSCV, 20 BD and 10 BD are used to quantize the coordinate and other attributes. We use 16:04 to split the coordinate map. HM18.0 \cite{HM} is used as a representation of standard video codecs in the experiments; additional selections of other standard video codecs are reported in section \ref{sec:ab_codec}. When using HM18.0, we set the maximum CU size to 64. 


\subsection{Experimental Results}
In this section, we report the performance of GSCV++ on both intra- and inter-prediction modes. For intra-prediction, we illustrate the RD curve of the first frame of each sequence as a representative example; for inter-prediction, we use the averaged frame-wise RD curve results within one GoP.

\subsubsection{Intra-Prediction Results}
Fig. \ref{fig:mian_intra} illustrates the RD curves of the proposed GSCV++, GSCV, GSCodec Studio, and GPCC v1. We see that: 1) GSCV++ reports significant gains compared to all the other methods; 2) Considering that GSCV++, GSCV, and GSCodec Studio are all video-based methods, the performance of GSCV++ is the best, followed by GSCV and GSCodec Studio; 3) GPCC v1 reports better performance in the low-bitstream range on the MPEG dataset because GPCC v1 is more effective at coordinate compression, and the coordinates occupy most of the bitstream in the low-bitstream region; 4) However, GPCC v1 reports different performance on the MPEG and AVS datasets: GPCC v1 performs better on the MPEG dataset but worse on the AVS dataset. After checking the dataset, we find that the primitive density of the AVS dataset is sparser than that of the MPEG sequences. Considering that GPCC v1 is initially designed for point clouds and is more powerful for compressing dense point clouds, sparse GS samples present a relatively challenging data distribution for GPCC v1; 5) Table \ref{tab:RD-gain} left part quantifies the averaged RD gains of GSCV++ on both the MPEG dataset and the AVS dataset. Specifically, GSCV++ exhibits gains of around 25\% to 40\% compared with GSCV. 


\begin{table}[h] \caption{RD gain of GSCV++ over GSCV}
\label{tab:RD-gain} \centering \setlength{\tabcolsep}{5pt} \begin{tabular}{|c|cccc|}
\hline
\multicolumn{1}{|l|}{} & \multicolumn{4}{c|}{Anchor: GSCV}                             \\ \hline
                       & \multicolumn{2}{c|}{Intra} & \multicolumn{2}{c|}{Inter} \\ \hline
Method         & \multicolumn{1}{c|}{MPEG}     & \multicolumn{1}{c|}{AVS}       & \multicolumn{1}{c|}{MPEG}      & AVS        \\ \hline
GSCV++           & \multicolumn{1}{c|}{-39.61\%} & \multicolumn{1}{c|}{-24.50\%}  & \multicolumn{1}{c|}{-68.70\%}  & -109.93\%  \\ \hline

\end{tabular} \end{table}



\begin{figure*}[t]
\centering
\begin{minipage}{0.24\textwidth}
    \includegraphics[width=\linewidth]{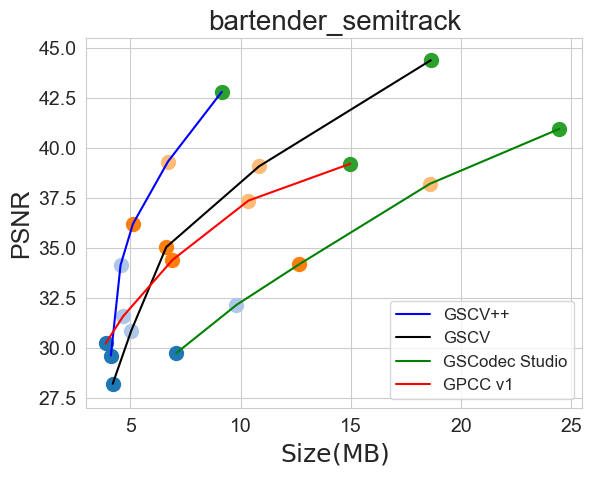}
\end{minipage}
\begin{minipage}{0.24\textwidth}
    \includegraphics[width=\linewidth]{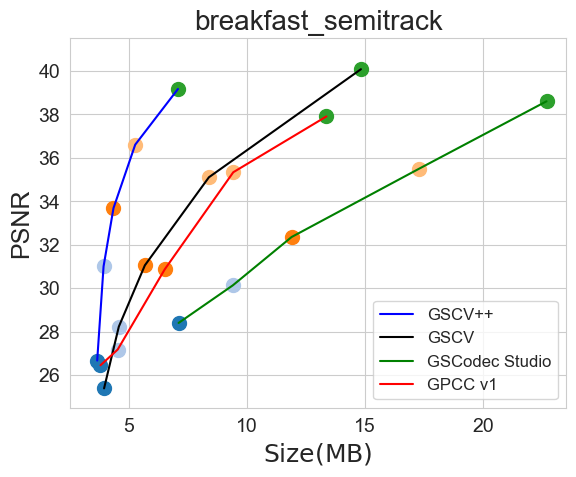}
\end{minipage}
\begin{minipage}{0.24\textwidth}
    \includegraphics[width=\linewidth]{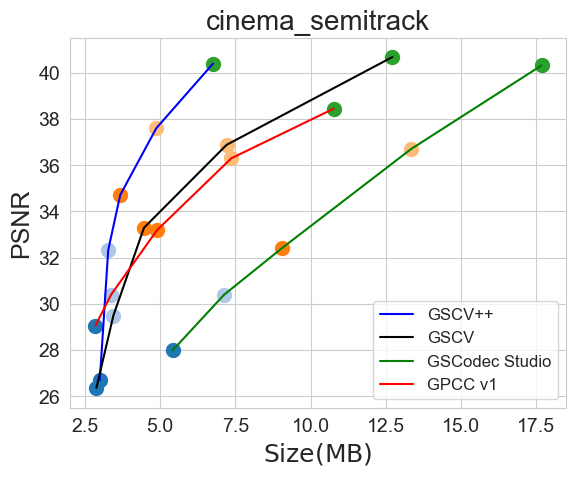}
\end{minipage}
\begin{minipage}{0.24\textwidth}
    \includegraphics[width=\linewidth]{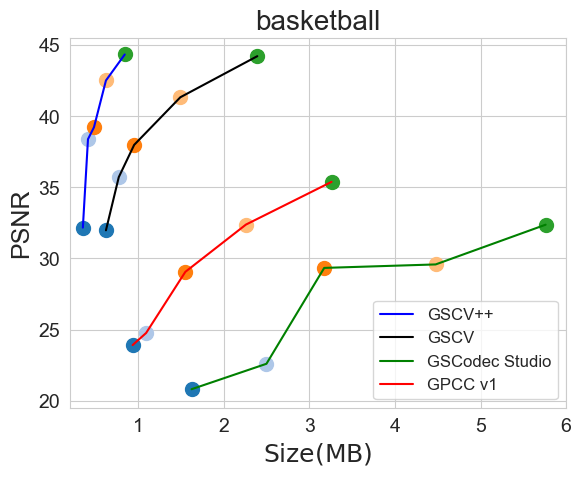}
\end{minipage}\\
\begin{minipage}{0.24\textwidth}
    \includegraphics[width=\linewidth]{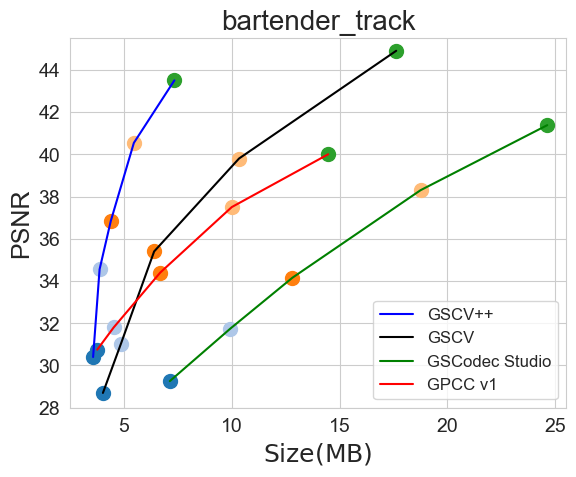}
\end{minipage}
\begin{minipage}{0.24\textwidth}
    \includegraphics[width=\linewidth]{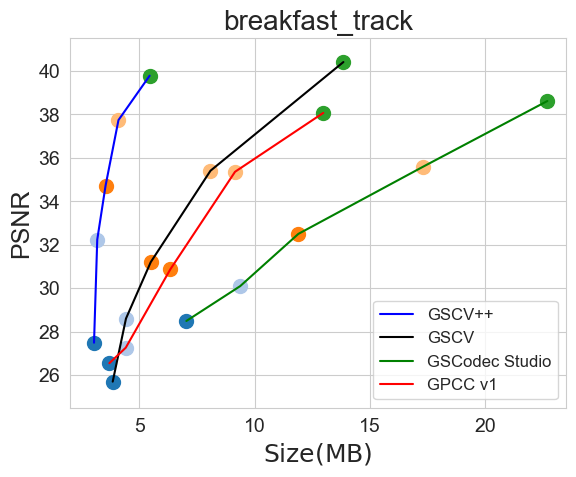}
\end{minipage}
\begin{minipage}{0.24\textwidth}
    \includegraphics[width=\linewidth]{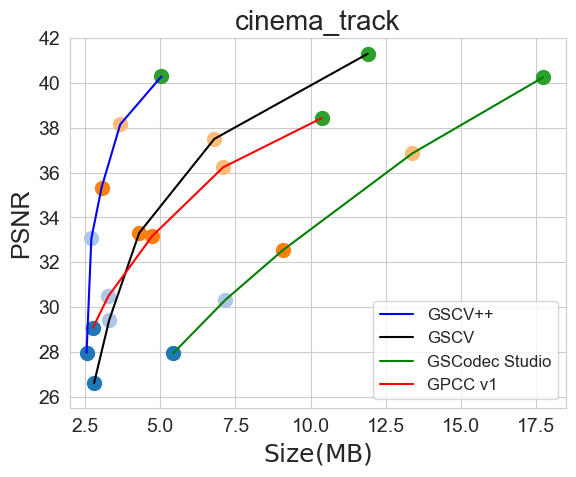}
\end{minipage}
\begin{minipage}{0.24\textwidth}
    \includegraphics[width=\linewidth]{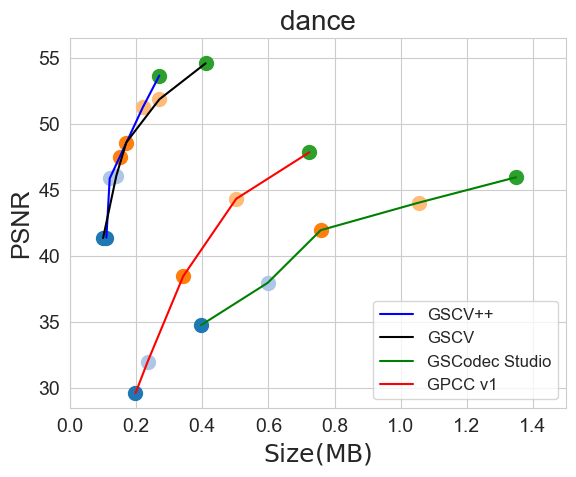}
\end{minipage}
\caption{Inter-prediction performance comparison on the MPEG and AVS datasets.}
\label{fig:mian_inter}
\end{figure*}

\subsubsection{Inter-Prediction Results}
Fig. \ref{fig:mian_inter} illustrates the RD curves of the proposed GSCV++, GSCV, GSCodec Studio, and GPCC v1 under the inter-prediction condition. We see that: 1) compared with the all-intra mode, GSCV++ demonstrates more pronounced gains under the inter-frame prediction mode; 2) different from the intra-prediction mode, where GSCodec Studio reports performance similar to GPCC v1 in most sequences; however, GSCodec Studio illustrates extremely poor performance under the inter-prediction mode. The reason is that GSCodec Studio uses the I-frame PLAS sorting index to generate P-frame 2D maps, which assumes that I- and P-frames are associated with stable tracking information. Without tracked information, the P-frame 2D maps of GSCodec Studio show significant content variations. It also reveals the importance of the proposed P-MiniPLAS module, which can construct robust tracked information for arbitrary GS sequences; 3) GSCV reports stable performance due to the introduction of inter-PLAS. However, GSCV++ includes more effective 2D map generation methods, resulting in even higher RD gains than those of the intra-prediction mode, as illustrated in Table \ref{tab:RD-gain} right part: the gains of GSCV++ compared to GSCV increased from around 25\% - 40\% to around 65\% - 110\%.

\subsubsection{Rate Allocation}\label{sec:exp_rate-allo}
\begin{figure}[!t]
\centering
\begin{minipage}{0.48\textwidth}
    \includegraphics[width=\linewidth]{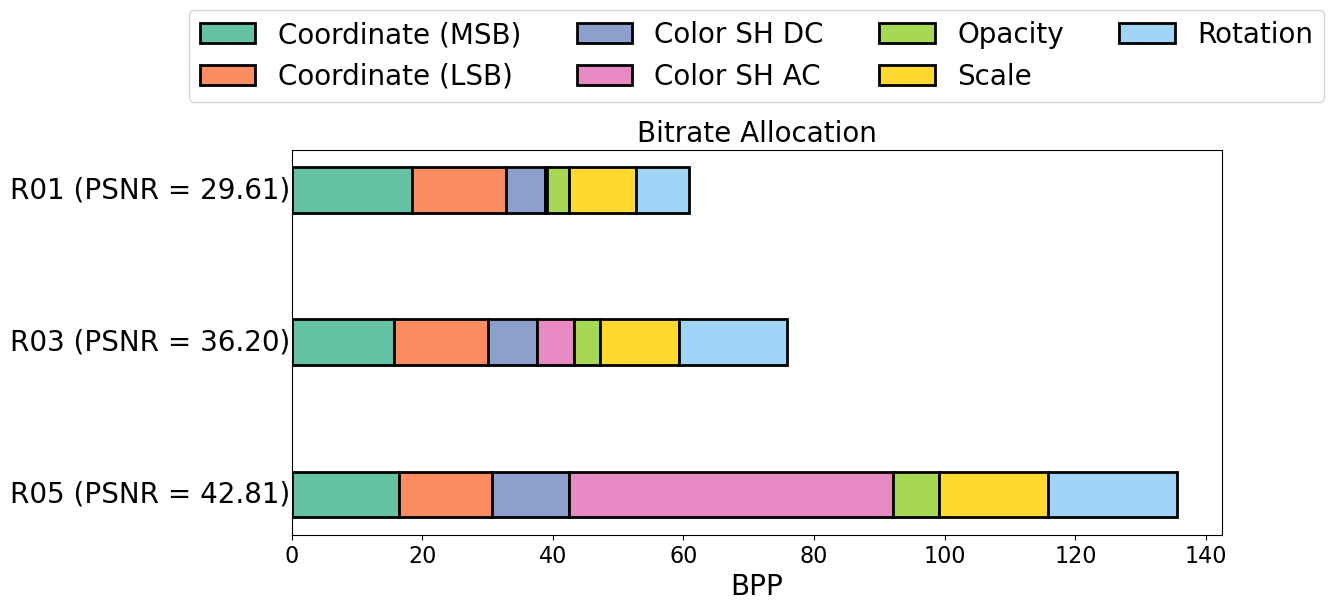}
\end{minipage}
\caption{Rate allocation of semi-tracked ``bartender''.}
\label{fig:rate}
\end{figure}

The rate allocation of ``bartender'' is shown in Fig. \ref{fig:rate}. We use the three rate points and detail the bits per primitive (BPP) for all the components. We observe that for the high-bitrate point, e.g., R05, the color SH AC occupies the largest part of the bitstream, while for the low-bitrate point, the coordinates account for nearly half of the total bitstream, e.g., over 30 BPP for coordinates. This reveals why video-based GS compression methods are worse than GPCC v1 for the low-bitrate region: GPCC v1 is more powerful for coordinate compression, enabling it to achieve around 10-15 coordinate BPP.

\subsubsection{Time Complexity}

In video-based GS compression frameworks, the computational cost mainly consists of two parts: 1) T1: the time spent converting vanilla GS into 2D video; and 2) T2: the subsequent video encoding and decoding time. First, Table \ref{tab:time} illustrates the T1 time of GSCV++ and GSCV on ``bartender''. Specifically, for GSCV++, there are three steps to obtain the final 2D GS videos: EA-SHDR, two-stage Morton scan, and P-MiniPLAS. For GSCV, there are two steps: spatial-stable initialization (SSI) and anchor-based PLAS refinement (A-PLASre). We see that for both I- and P-frames, GSCV++ can exhibit a faster 2D map generation speed, which can benefit the practical application of 3D GS. 
\begin{table}[!t] \caption{Computation time (s) of preprocessing and 2D map generation for GSCV++ and GSCV}
\label{tab:time} \centering \resizebox{\columnwidth}{!}{
\begin{tabular}{|c|cccc|ccc|}
\hline
Method  & \multicolumn{4}{c|}{GSCV++}                                                                  & \multicolumn{3}{c|}{GSCV}                                      \\ \hline
Type &
  \multicolumn{1}{c|}{EA-SHDR} &
  \multicolumn{1}{c|}{\makecell{Two-stage\\Morton Scan}} &
  \multicolumn{1}{c|}{\makecell{P-MiniPLAS\\(MBS = 64)}} &
  All &
  \multicolumn{1}{c|}{SSI} &
  \multicolumn{1}{c|}{Inter-PLAS} &
  All \\ \hline
I-frame & \multicolumn{1}{c|}{0.54} & \multicolumn{1}{c|}{0.075} & \multicolumn{1}{c|}{3.96}  & \textbf{4.575}  & \multicolumn{1}{c|}{0.07} & \multicolumn{1}{c|}{24.71} & 24.78 \\ \hline
P-frame & \multicolumn{1}{c|}{0.05} & \multicolumn{1}{c|}{0.075} & \multicolumn{1}{c|}{14.54} & \textbf{14.665} & \multicolumn{1}{c|}{0.09} & \multicolumn{1}{c|}{74.51} & 74.6  \\ \hline
\end{tabular}
} \end{table}

Second, the T2 time is related to the downstream codecs used in practical compression. Therefore, we list the encoding time (En) and decoding time (De) of GSCV++, GSCV, GSCodec Studio (GSCS), and GPCC v1 in Table \ref{tab:codec_time}. For GSCV++ and GSCV, we test two different video codecs, HM18.0 and FFMPEG libx265. We see that: 1) GSCV++ libx265 reports the fastest encoding and decoding times; 2) using the same video codecs, GSCV++ requires less encoding and decoding time than GSCV and GSCodec Studio. It indicates that the generated GS maps are more user-friendly for standard video codecs by introducing smoother textures; 3) using GSCV++ libx265 R05 as an example, the average encoding and decoding times per image are around $2.39/14 = 0.171s$ and $0.50/14 = 0.035s$. This means that currently, GSCV++ can achieve almost real-time decoding when employing multi-threaded parallel decoding (around 30 FPS), considering that the decoding process of different GS attributes is independent. Given that GS cannot currently be generated in real time, real-time encoding is not a strict industrial requirement. Even so, reducing encoding latency remains beneficial for facilitating broader GS deployment. One promising method is to control the number of primitives before compression, such as introducing pruning during GS generation \cite{yanghybridgs} or adding it as part of the preprocessing step.

\begin{table}[!t] \caption{The averaged frame-wise encoding and decoding times of different methods on the tracked ``bartender''.}
\label{tab:codec_time} \centering
\scalebox{0.7}{
\begin{tabular}{|c|cc|cc|cc|cc|cc|cc|}
\hline
Codec &
\multicolumn{2}{c|}{\makecell{GSCV++\\HM18.0}} &
 \multicolumn{2}{c|}{\makecell{GSCV++\\libx265}} &
  \multicolumn{2}{c|}{\makecell{GSCV\\HM18.0}} &
   \multicolumn{2}{c|}{\makecell{GSCV\\libx265}} &
   \multicolumn{2}{c|}{\makecell{GSCS\\libx265}} &
  \multicolumn{2}{c|}{\makecell{GPCC\\v1}} \\ \hline
\multicolumn{1}{|c|}{\makecell{Rate\\Point}} &
  \multicolumn{1}{c|}{En} &
  De &
  \multicolumn{1}{c|}{En} &
  De &
  \multicolumn{1}{c|}{En} &
  De &
  \multicolumn{1}{c|}{En} &
  De &
  \multicolumn{1}{c|}{En} &
  De &
  \multicolumn{1}{c|}{En} &
  De \\ \hline
R05 &
  \multicolumn{1}{c|}{397.8} &
  1.4 &
  \multicolumn{1}{c|}{2.4} &
  0.5 &
  \multicolumn{1}{c|}{1183.5} &
  3.9 &
  \multicolumn{1}{c|}{4.6} &
  1.2 &
  \multicolumn{1}{c|}{8.9} &
  1.4 &
  \multicolumn{1}{c|}{304.1} &
  86.0 \\ \hline
R04 &
  \multicolumn{1}{c|}{332.1} &
  1.2 &
  \multicolumn{1}{c|}{2.4} &
  0.4 &
  \multicolumn{1}{c|}{1012.5} &
  2.9 &
  \multicolumn{1}{c|}{4.1} &
  0.9 &
  \multicolumn{1}{c|}{8.5} &
  1.4 &
  \multicolumn{1}{c|}{298.0} &
  88.4 \\ \hline
R03 &
  \multicolumn{1}{c|}{246.6} &
  1.0 &
  \multicolumn{1}{c|}{2.2} &
  0.4 &
  \multicolumn{1}{c|}{540.3} &
  2.1 &
  \multicolumn{1}{c|}{2.8} &
  0.7 &
  \multicolumn{1}{c|}{7.3} &
  1.3 &
  \multicolumn{1}{c|}{291.7} &
  86.7 \\ \hline
R02 &
  \multicolumn{1}{c|}{213.1} &
  0.9 &
  \multicolumn{1}{c|}{2.2} &
  0.4 &
  \multicolumn{1}{c|}{353.4} &
  1.6 &
  \multicolumn{1}{c|}{2.3} &
  0.6 &
  \multicolumn{1}{c|}{6.4} &
  1.3 &
  \multicolumn{1}{c|}{291.9} &
  86.0 \\ \hline
R01 &
  \multicolumn{1}{c|}{187.3} &
  0.8 &
  \multicolumn{1}{c|}{2.1} &
  0.3 &
  \multicolumn{1}{c|}{301.5} &
  1.3 &
  \multicolumn{1}{c|}{1.9} &
  0.5 &
  \multicolumn{1}{c|}{5.2} &
  1.1 &
  \multicolumn{1}{c|}{293.2} &
  85.4 \\ \hline

\end{tabular}
} \end{table}

\subsection{Ablation Study}
In this section, we report the ablation study of GSCV++. Specifically, we use three MPEG semi-tracked sequences as test sequences. If the ablation experiments exhibit consistent performance across the three sequences, we report the average results over all three. Otherwise, when the performance is inconsistent, we select the sequences with divergent outcomes for further analysis and discussion.

\subsubsection{Effectiveness of EA-SHDR and P-MiniPLAS}

To quantify the contribution of the two modules, we evaluate four combinations of EA-SHDR and P-MiniPLAS, Table~\ref{tab:module_cumulative} reports the arithmetic mean of the three per-sequence rate changes, using the setting with both modules disabled as the common reference. Enabling EA-SHDR alone reports a 21.92\% RD gain, while enabling P-frame refinement alone illustrates 32.43\% gain. Combining both modules yields the largest average gain, 41.79\%, and also provides the largest reduction relative to the common baseline on each sequence. Using the EA-SHDR-only setting as the reference, adding P-frame refinement provides a further average gain of 24.65\%. Conversely, adding EA-SHDR to the refinement-only setting yields a gain of 10.67\%. These results support the complementary roles of the two modules in the main configuration.

\begin{table}[h]
\centering
\caption{Average RD changes of module combinations. The last row is the main GSCV++ configuration.}
\label{tab:module_cumulative}
\small
\setlength{\tabcolsep}{7pt}
\begin{tabular}{ccc}
\hline
EA-SHDR & P-MiniPLAS & Avg. Gain (\%) \\
\hline
w/o & w/o & 0.00 \\
w & w/o & -21.92 \\
w/o & w & -32.43 \\
\textbf{w} & \textbf{w} & \textbf{-41.79} \\
\hline
\end{tabular}
\par\smallskip
\begin{minipage}{\columnwidth}
\end{minipage}
\end{table}

\begin{figure}[!t]
\centering
\begin{minipage}{0.24\textwidth}
    \includegraphics[width=\linewidth]{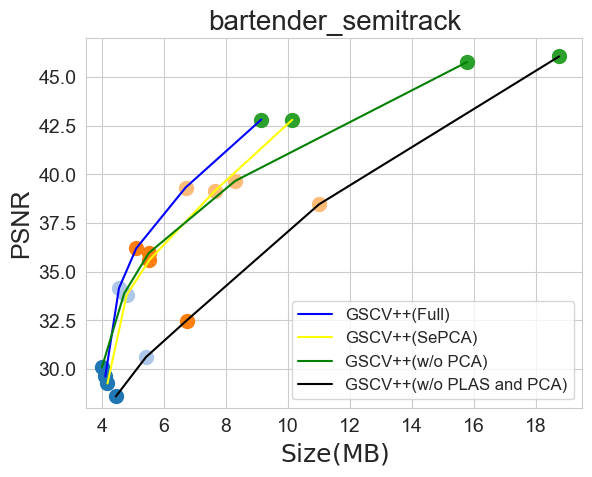}
\end{minipage}
\begin{minipage}{0.24\textwidth}
    \includegraphics[width=\linewidth]{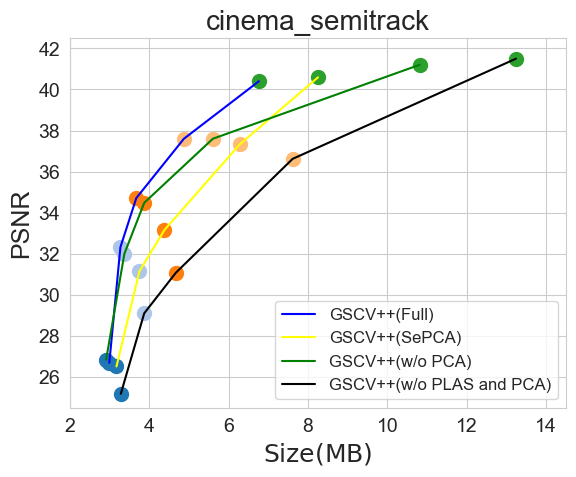}
\end{minipage}
\caption{Ablation Study of EA-SHDR and P-MiniPLAS on the MPEG semi-tracked dataset.}
\label{fig:ab_module}
\end{figure}

To better illustrate the influence of different modules on different bitrate ranges, we additionally present EA-SHDR and P-MiniPLAS in RD curves using three alternatives: 1) disable both modules (w/o PLAS and PCA); 2) disable EA-SHDR (w/o PCA); and 3) use separate PCA bases for I- and P-frames (SePCA). We show the results of ``bartender'' and ``cinema'' in Fig. \ref{fig:ab_module}. We see that both EA-SHDR and P-MiniPLAS contribute to the overall performance. Introducing PCA-related operations will incur a lower-quality upper bound; however, for ``bartender'', the performance of SePCA and w/o PCA is close, while the performance without PCA is superior to SePCA for ``cinema''. This indicates that the principal component space of ``cinema'' color SH AC exhibits significant variation across different frames. Using I-frame truncated SVD results for P-frames is important to ensure stable prediction gain.

We further compare the complete Inter-PLAS \cite{GSCV} with our P-MiniPLAS on the same sequences, retaining EA-SHDR, shared I/P PCA bases, coordinate coding, HM settings, and the five operating points. The Inter-PLAS alternative applies vanilla PLAS to the I-frame and globally refines each P-frame using its own I-frame ordering and quantized anchor, with unit attribute weights, radius decay 0.95, and minimum sorting block size 16. The results are shown in Table~\ref{tab:interplas_workflow}. The proposed P-MiniPLAS reports an obvious gain of 24.49\% on average, with gains of 24.11\%, 28.87\%, and 20.50\% on ``bartender'', ``breakfast'', and ``cinema'', respectively. Mean 2D map generation time decreases from 25.61 to 2.10 s for the I-frame and from 63.04 to 16.43 s per P-frame, corresponding to a 3.84$\times$ P-frame sorting speedup. These results show consistent RD improvements and lower sorting cost across the three sequences.

\begin{table}[h]
\centering
\caption{Average RD change and computation time of complete I/P map generation between Inter-PLAS and P-MiniPLAS.}
\label{tab:interplas_workflow}
\small
\setlength{\tabcolsep}{4pt}
\begin{tabular}{lccc}
\hline
Method & \shortstack{$\Delta R$\\(\%)} & \shortstack{I sorting\\(s)} & \shortstack{P sorting\\(s/frame)} \\
\hline
Inter-PLAS & 0.00 & 25.61 & 63.04 \\
\textbf{P-MiniPLAS} & \textbf{-24.49} & \textbf{2.10} & \textbf{16.43} \\
\hline
\end{tabular}
\par\smallskip
\begin{minipage}{\columnwidth}
\end{minipage}
\end{table}

\subsubsection{Energy Ratio for EA-SHDR}

\begin{figure}[!t]
\centering
\begin{minipage}{0.24\textwidth}
    \includegraphics[width=\linewidth]{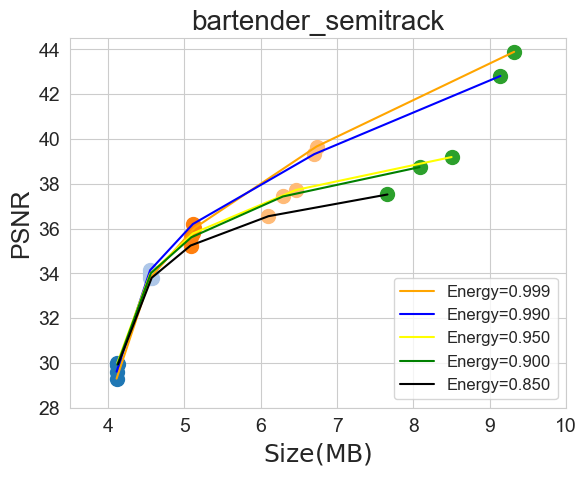}
\end{minipage}
\begin{minipage}{0.24\textwidth}
    \includegraphics[width=\linewidth]{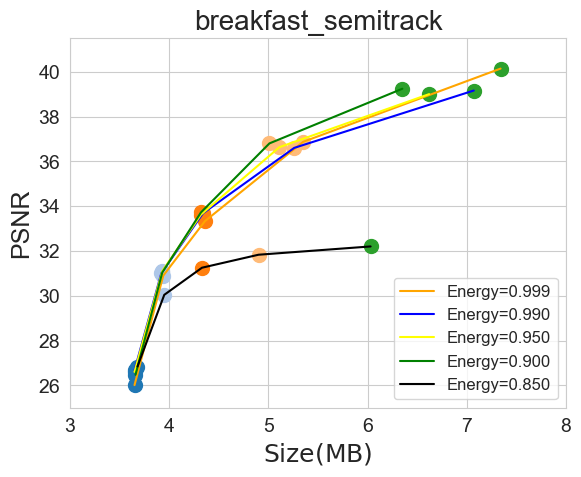}
\end{minipage}
\caption{Ablation Study of EA-SHDR energy ratio on the MPEG semitracked dataset.}
\label{fig:ab_pca}
\end{figure}

In Section \ref{sec:exp_mp}, we set the energy ratio as $A = 0.99 $. In this part, we show the results of $A = 0.999,\ 0.950,\ 0.900,\ 0.980$ in Fig. \ref{fig:ab_pca}. We see that the ``bartender'' and ``breakfast'' report different RD curves corresponding to a monotonic variation of energy, while ``cinema'' is close to ``breakfast''. For ``bartender'', it shows perfect hierarchical performance with the increase of energy,  while ``breakfast'' reports the best RD results when $A = 0.900$ and an obvious performance gap with $A = 0.850$. The reason is that the texture of ``breakfast'' is more complex than that of ``bartender'', and the energy distribution is different: the latent feature channels $k^*$ for ``bartender'' and ``breakfast'' are [27, 21, 15, 12, 9] and [24, 18, 12, 9, 6], respectively. The energy of ``breakfast'' is more concentrated. The best RD curve corresponds to finding the best solution of $J = D + \lambda R$, and ``bartender'' has different global optima from ``breakfast''. Further work is needed to explore an RD analysis for the GS 2D map corresponding to the color SH AC energy.

\subsubsection{Influence of PLAS Weighting Factor}\label{sec:ab_plasweight}
\begin{figure}[!t]
\centering
\subfloat[PLAS Weighting Factors]{\begin{minipage}{0.24\textwidth}\includegraphics[width=\linewidth]{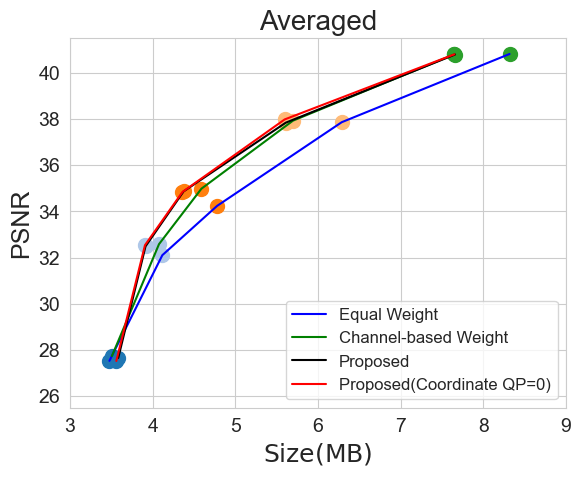}\end{minipage}}
\subfloat[CU Size]{\begin{minipage}{0.24\textwidth}\includegraphics[width=\linewidth]{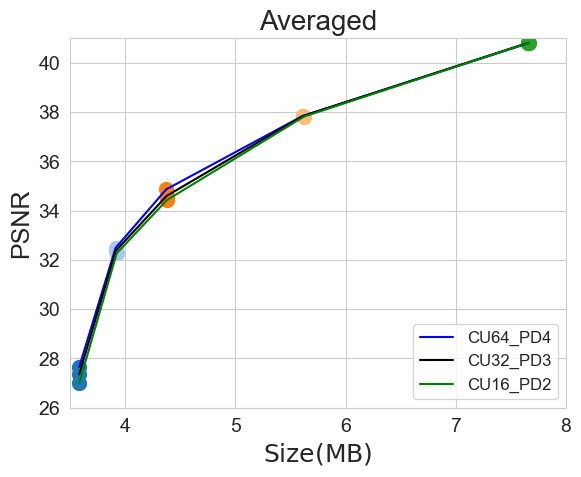}\end{minipage}}
\caption{(a): Ablation study of PLAS weight; (b): Ablation study of CU size. }
\label{fig:ab_plasweight}
\end{figure}

In Section \ref{sec:plasWeight}, we introduce a prior-based PLAS weighting factor to realize content-aware PLAS image smoothness. To highlight the effectiveness of this method, we report the performance of GSCV++ corresponding to: 1) equal weighting factors for all the GS attributes; 2) attribute channel-based weighting factors, i.e., coordinate: color SH DC: color SH AC: opacity: scale: rotation = 3:3:45:1:3:4; and 3) when calculating the weight of coordinate, we fix $QP = 0$ for $f^2(QP^a)$ (noted as proposed (coordinate QP = 0)), which means that regardless of the QP range of other attributes, the weighting factors of coordinate are fixed.  The results are shown in Fig. \ref{fig:ab_plasweight} (a). We observe that the performance of the proposed and proposed (coordinate QP = 0) are the best, followed by channel-based and equal weight. Using equal weight as a baseline, the RD gains of the channel-based, proposed, and proposed (coordinate QP = 0) methods are 6.2\%, 7.5\%, and 8.3\%, respectively. Using coordinate QP as 0 is slightly better than setting coordinate $QP = QP_{DC}$.  It is worth noting that the meaning of QP=0 differs across different encoders: nearly lossless for HM/VTM, while real lossless for FFMPEG libx264. Considering that the proposed method requires some prior knowledge, such as the QP of the downstream video codecs, we suggest using channel-based weights when QP prior knowledge is not available.


\subsubsection{Influence of CU Size}\label{sec:exp_ab_cusize}
In Section \ref{sec:inter-miniplas}, we introduce the setting of the P-frame MBS, which can be used as a prior for CU size selection in the downstream video codec. In Section \ref{sec:exp_mp}, we set MBS to 64 and adopt a progressive block size of [64x64, 32x32, 16x16, 8x8, 4x4] for Inter-MiniPLAS. In this Section, we report the performance of using maximum CU size and partition depth (PD) as  (64, 4), (32, 3), and (16,2) for real coding, ensuring that the minimum CU size is 8x8. The results are shown in Fig. \ref{fig:ab_plasweight} (b). We see that the performance of different CU and PD settings yields similar RD performance: using CU16\_PD2 as a benchmark, the performance gains of CU32\_PD3 and CU64\_PD4 are 0.33\% and 0.41\%, respectively. 

One advantage of using a smaller CU size and PD is low encoding complexity. According to the RDO algorithm, the encoder traverses all possible partitioning schemes and computes the reconstruction error and bitrate to determine the optimal choice. Smaller CU involves fewer partitioning schemes, allowing the optimal coding decision to be reached more efficiently. Using the ``cinema'' inter-prediction mode with the highest bitrate as an example, the averaged HM18.0 encoding times of each 2D GS map of CU16\_PD2, CU32\_PD3, and CU64\_PD4 are 16.39, 18.85, and 22.45. For low-delay applications, P-MiniPLAS can provide an important prior for the trade-off between RD and encoding time.

\subsubsection{Influence of MBS}
\begin{figure}[!t]
\centering
\subfloat[P-MiniPLAS with different MBS and CU combination.]{\begin{minipage}{0.24\textwidth}\includegraphics[width=\linewidth]{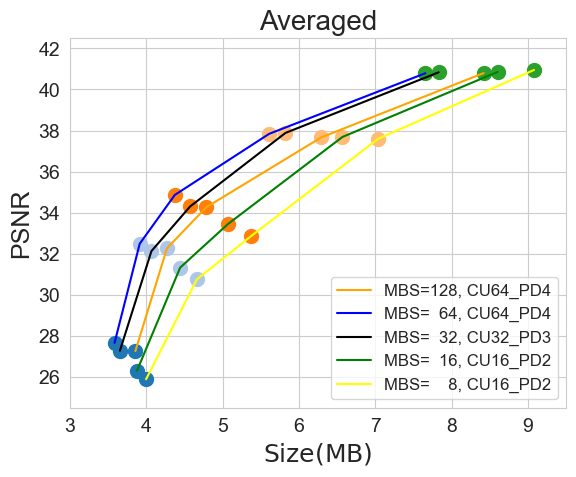}\end{minipage}}
\subfloat[Small MBS with larger CU size.]{\begin{minipage}{0.24\textwidth}\includegraphics[width=\linewidth]{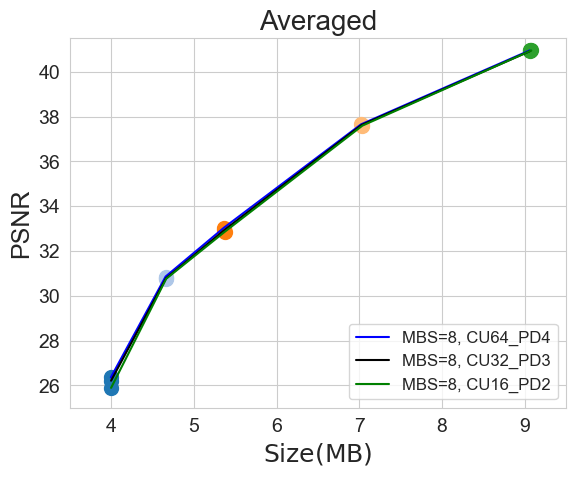}\end{minipage}}
\caption{(a): Ablation study of P-MiniPLAS MBS on the MPEG semi-tracked dataset; (b): Ablation study of CU size on small MBS setting. }
\label{fig:ab_PplasCU}
\end{figure}

We illustrate the performance of GSCV++ with different MBS in this part. First, we test the performance of different MBS, including MBS = 128, 64, 32, 16, and 8. Given the MBS, we configure the encoder by selecting the maximum CU size that is closest to the MBS. For example, MBS = 128 with CU size = 64, MBS = 32 with CU size = 32, and MBS = 8 with CU size = 16. The results are shown in Fig. \ref{fig:ab_PplasCU} (a). We observe that: 1) the overall performance increases from MBS = 8 to 64, but decreases when MBS = 128. This result indicates that larger MBS does not necessarily lead to better performance. Considering that P-MiniPLAS is a progressive process, an excessively large MBS can lead to local optima and prevent subsequent smaller block size values from escaping these local optima. The discussion about Fig. \ref{fig:dcvsxyz} can also explain this problem: the G1 attributes require a larger BPP with larger MBSs; 2) a larger MBS requires more computation time. Using ``bartender'' as an example, it has around half a million primitives, and the P-MiniPLAS times are 17.27s, 13.7s, 6.24s, 5.76s, 1.40s from MBS = 128 to 8. With a reasonable MBS value range, the compression ratio generally tends to increase proportionally with the P-MiniPLAS computation time.

Section \ref{sec:inter-miniplas} motivates limiting the maximum CU size using MBS as an engineering prior. To evaluate this design choice, we illustrate additional experiments with MBS = 8 and three different maximum CU sizes that are larger than 8 in Fig. \ref{fig:ab_PplasCU} (b). The three curves are observed to nearly overlap, which supports this observation on the tested sequences.

\subsubsection{Influence of Differentiated Encoding on Coordinate}\label{sec:diff_xyz_ablation}

Section \ref{sec:diff_xyz} introduces non-uniform coordinate splitting. Table \ref{tab:ab_xyz} compares lossless HM18.0 and LZMA for LSB coding on three semi-tracked sequences, using 10:10, 12:08, and 16:04 splits in all-intra and inter modes. The MSB uses HM18.0. LZMA receives direct LSB data for I-frames and I-frame-referenced residuals for subsequent frames. The highlighted entries indicate the smallest total coordinate size among the tested configurations for each sequence and mode.

The results support a mode- and split-dependent LSB choice. For all-intra coding, LZMA gives a smaller LSB size than HM18.0 at 12:08 and 16:04 on all three sequences; the 10:10 split instead favors HM18.0. For inter coding, HM18.0 gives a smaller LSB size for all nine tested sequence/split combinations. We therefore recommend LZMA for the evaluated all-intra configurations and HM18.0 for the evaluated inter configurations. 

\begin{table}[!t] \caption{Coordinate compression of MPEG semi-tracked sequences.}
\label{tab:ab_xyz} \centering
\scalebox{0.75}{
\begin{tabular}{|c|cc|ccc|ccc|}
\hline
 &
  \multicolumn{2}{c|}{} &
  \multicolumn{3}{c|}{\textbf{MSB}:HM18.0 \textbf{LSB}:HM18.0} &
  \multicolumn{3}{c|}{\textbf{MSB}:HM18.0 \textbf{LSB}:LZMA} \\ \cline{4-9} 
\multirow{-2}{*}{Sequence} &
  \multicolumn{2}{c|}{\multirow{-2}{*}{Setting}} &
  \multicolumn{1}{c|}{MSB} &
  \multicolumn{1}{c|}{LSB} &
  ALL &
  \multicolumn{1}{c|}{MSB} &
  \multicolumn{1}{c|}{LSB} &
  ALL \\ \hline
 &
  \multicolumn{1}{c|}{} &
  10:10 &
  \multicolumn{1}{c|}{1.99} &
  \multicolumn{1}{c|}{28.06} &
  30.05 &
  \multicolumn{1}{c|}{1.99} &
  \multicolumn{1}{c|}{30.88} &
  32.86 \\ \cline{3-3} 
 &
  \multicolumn{1}{c|}{} &
  12:08 &
  \multicolumn{1}{c|}{4.93} &
  \multicolumn{1}{c|}{25.04} &
  29.96 &
  \multicolumn{1}{c|}{4.93} &
  \multicolumn{1}{c|}{24.00} &
  28.92 \\ \cline{3-3} 
 &
  \multicolumn{1}{c|}{\multirow{-3}{*}{All-intra}} &
  16:04 &
  \multicolumn{1}{c|}{15.64} &
  \multicolumn{1}{c|}{13.62} &
  29.25 &
  \multicolumn{1}{c|}{15.64} &
  \multicolumn{1}{c|}{12.46} &
  {\color[HTML]{FF0000} 28.1} \\ \cline{2-9} 
 &
  \multicolumn{1}{c|}{} &
  10:10 &
  \multicolumn{1}{c|}{2.86} &
  \multicolumn{1}{c|}{27.56} &
  30.42 &
  \multicolumn{1}{c|}{2.86} &
  \multicolumn{1}{c|}{28.36} &
  31.22 \\ \cline{3-3} 
 &
  \multicolumn{1}{c|}{} &
  12:08 &
  \multicolumn{1}{c|}{5.72} &
  \multicolumn{1}{c|}{24.73} &
  30.45 &
  \multicolumn{1}{c|}{5.72} &
  \multicolumn{1}{c|}{25.43} &
  31.15 \\ \cline{3-3} 
\multirow{-6}{*}{bartender} &
  \multicolumn{1}{c|}{\multirow{-3}{*}{Inter}} &
  16:04 &
  \multicolumn{1}{c|}{16.41} &
  \multicolumn{1}{c|}{13.62} &
  {\color[HTML]{FF0000} 30.03} &
  \multicolumn{1}{c|}{16.41} &
  \multicolumn{1}{c|}{14.32} &
  30.73 \\ \hline
 &
  \multicolumn{1}{c|}{} &
  10:10 &
  \multicolumn{1}{c|}{2.27} &
  \multicolumn{1}{c|}{28.05} &
  30.32 &
  \multicolumn{1}{c|}{2.27} &
  \multicolumn{1}{c|}{30.97} &
  33.24 \\ \cline{3-3} 
 &
  \multicolumn{1}{c|}{} &
  12:08 &
  \multicolumn{1}{c|}{5.36} &
  \multicolumn{1}{c|}{24.87} &
  30.23 &
  \multicolumn{1}{c|}{5.36} &
  \multicolumn{1}{c|}{24.00} &
  29.36 \\ \cline{3-3} 
 &
  \multicolumn{1}{c|}{\multirow{-3}{*}{All-intra}} &
  16:04 &
  \multicolumn{1}{c|}{16.07} &
  \multicolumn{1}{c|}{13.62} &
  29.69 &
  \multicolumn{1}{c|}{16.07} &
  \multicolumn{1}{c|}{12.46} &
  {\color[HTML]{FF0000} 28.53} \\ \cline{2-9} 
 &
  \multicolumn{1}{c|}{} &
  10:10 &
  \multicolumn{1}{c|}{3.14} &
  \multicolumn{1}{c|}{26.93} &
  30.08 &
  \multicolumn{1}{c|}{3.14} &
  \multicolumn{1}{c|}{28.14} &
  31.29 \\ \cline{3-3} 
 &
  \multicolumn{1}{c|}{} &
  12:08 &
  \multicolumn{1}{c|}{6.02} &
  \multicolumn{1}{c|}{23.92} &
  29.93 &
  \multicolumn{1}{c|}{6.02} &
  \multicolumn{1}{c|}{25.03} &
  31.05 \\ \cline{3-3} 
\multirow{-6}{*}{cinema} &
  \multicolumn{1}{c|}{\multirow{-3}{*}{Inter}} &
  16:04 &
  \multicolumn{1}{c|}{15.88} &
  \multicolumn{1}{c|}{13.62} &
  {\color[HTML]{FF0000} 29.5} &
  \multicolumn{1}{c|}{15.88} &
  \multicolumn{1}{c|}{14.16} &
  30.05 \\ \hline
 &
  \multicolumn{1}{c|}{} &
  10:10 &
  \multicolumn{1}{c|}{5.44} &
  \multicolumn{1}{c|}{30.77} &
  36.22 &
  \multicolumn{1}{c|}{5.44} &
  \multicolumn{1}{c|}{31.85} &
  37.29 \\ \cline{3-3} 
 &
  \multicolumn{1}{c|}{} &
  12:08 &
  \multicolumn{1}{c|}{10.22} &
  \multicolumn{1}{c|}{25.52} &
  35.74 &
  \multicolumn{1}{c|}{10.22} &
  \multicolumn{1}{c|}{24.00} &
  {\color[HTML]{FF0000} 34.22} \\ \cline{3-3} 
 &
  \multicolumn{1}{c|}{\multirow{-3}{*}{All-intra}} &
  16:04 &
  \multicolumn{1}{c|}{21.99} &
  \multicolumn{1}{c|}{13.62} &
  35.61 &
  \multicolumn{1}{c|}{21.99} &
  \multicolumn{1}{c|}{12.48} &
  34.47 \\ \cline{2-9} 
 &
  \multicolumn{1}{c|}{} &
  10:10 &
  \multicolumn{1}{c|}{4.82} &
  \multicolumn{1}{c|}{27.91} &
  32.72 &
  \multicolumn{1}{c|}{4.82} &
  \multicolumn{1}{c|}{28.28} &
  33.10 \\ \cline{3-3} 
 &
  \multicolumn{1}{c|}{} &
  12:08 &
  \multicolumn{1}{c|}{7.93} &
  \multicolumn{1}{c|}{24.44} &
  {\color[HTML]{FF0000} 32.38} &
  \multicolumn{1}{c|}{7.93} &
  \multicolumn{1}{c|}{25.18} &
  33.11 \\ \cline{3-3} 
\multirow{-6}{*}{breakfast} &
  \multicolumn{1}{c|}{\multirow{-3}{*}{Inter}} &
  16:04 &
  \multicolumn{1}{c|}{18.9} &
  \multicolumn{1}{c|}{13.62} &
  32.52 &
  \multicolumn{1}{c|}{18.9} &
  \multicolumn{1}{c|}{14.29} &
  33.20 \\ \hline
\end{tabular}
} \end{table}

\subsubsection{Performance of Different Standard Codecs}\label{sec:ab_codec}

\begin{figure}[!t]
\centering
\begin{minipage}{0.45\textwidth}
    \includegraphics[width=\linewidth]{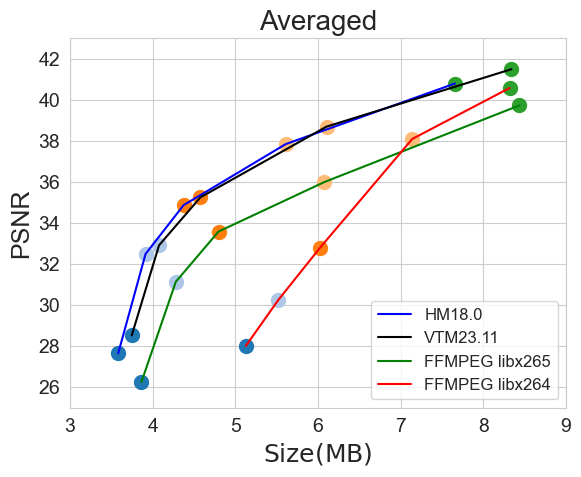}
\end{minipage}
\caption{Performance of GSCV++ with different video codecs.}
\label{fig:ab_condecs}
\end{figure}

In Section \ref{sec:exp_mp}, we use HM18.0 as the representation of standard video codecs for GSCV++. In Fig. \ref{fig:ab_condecs}, we report the performance of using VTM23.11 \cite{VTM}, FFMPEG libx265, and FFMPEG libx264 \cite{ffmpeg} as practical codecs. Specifically, the libx265 and libx264 configurations evaluated here support maximum input BDs of 12 and 10, respectively. Therefore, we use 12:08 for coordinate compression for FFMPEG libx265, while 10:10 for FFMPEG libx264. We see that: 1) the performance of HM18.0 and VTM23.11 is close, followed by FFMPEG libx265 and libx264. Using FFMPEG libx264 as the benchmark, the RD gains of libx265, VTM23.11, and HM18.0 are 12.01\%, 23.27\%, and 23.91\%; 2) VTM23.11 does not report better performance than HM18.0, which conflicts with the conclusion on natural videos. The reason is that compared with natural images, GS 2D maps illustrate a more pronounced heavy-tail behavior, and these maps do not provide any explicit motion like natural videos. Some advanced video coding modules involved in VVC do not illustrate significant gains over HEVC.

\begin{figure}[!t]
\centering
\subfloat[Residual histogram of rendered-view images. ]{\begin{minipage}{0.45\textwidth}\includegraphics[width=\linewidth]{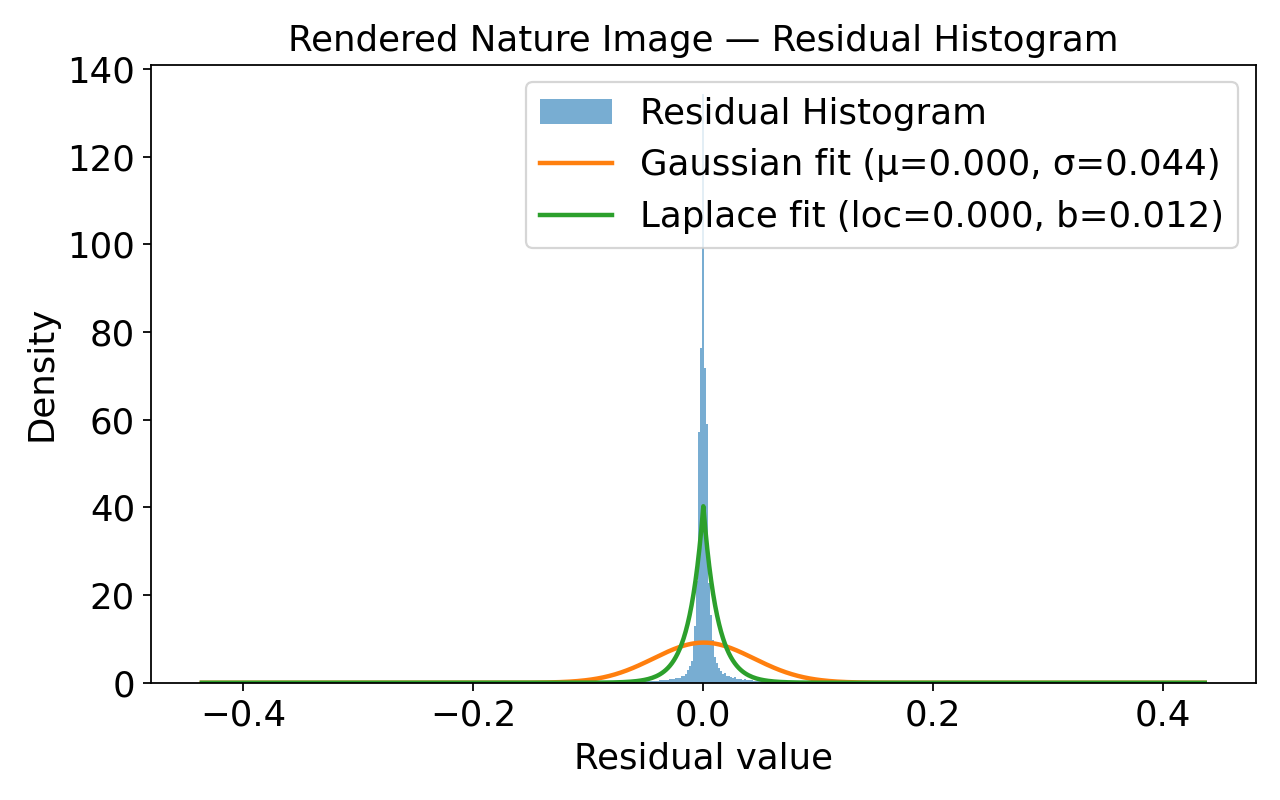}\end{minipage}}\\
\subfloat[Residual histogram of P-MiniPLAS scale map.]{\begin{minipage}{0.45\textwidth}\includegraphics[width=\linewidth]{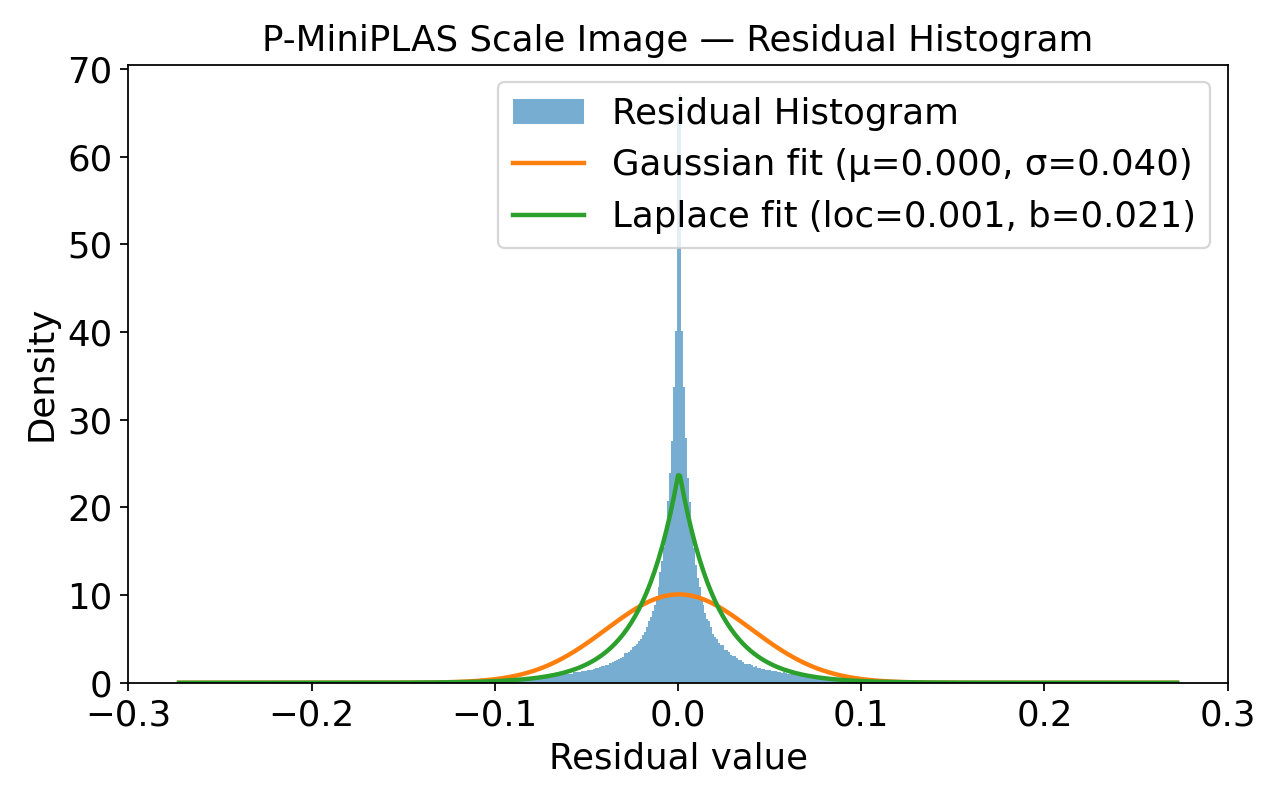}\end{minipage}}
\caption{Residual Analysis. }
\label{fig:ab_res}
\end{figure}

We use the ``cinema'' rendered view 9th images from frames 0 and 1 as the representations of natural images. The residual histogram for the inter-prediction is demonstrated in Fig. \ref{fig:ab_res}(a), and the same residual analysis of P-MiniPLAS scale maps is used for comparison and illustrated in Fig. \ref{fig:ab_res}(b). Based on the histogram, we also provide fitted Gaussian and Laplace curves. The larger Laplace scale parameter $b$ of the natural image is 0.012, while that of the scale map is 0.021. This reflects the conclusion that P-MiniPLAS images demonstrate an obvious heavy-tail behavior. Additionally, the entropy of the natural and scale residual images is 4.26 and 6.44, respectively. This evidence reveals that the P-MiniPLAS image presents a more challenging condition for current standard video codecs.

\subsubsection{Performance of Different GoP Size}\label{sec:ab_gop}

To evaluate the effect of GoP size beyond the eight-frame setting, we compress the same 32 consecutive frames of the MPEG semi-tracked ``cinema'' sequence using GoP sizes of 8, 16, and 32. These settings partition the sequence into four, two, and one independently coded groups, respectively, each containing one I-frame followed by P-frames that reference that group's I-frame. We retain the main parameters and five operating points. Each new I-frame refreshes its 2D mapping and EA-SHDR basis.

\begin{figure}[!t]
\centering
\subfloat[RD performance on the same 32 frames.]{\begin{minipage}{0.45\textwidth}\includegraphics[width=\linewidth]{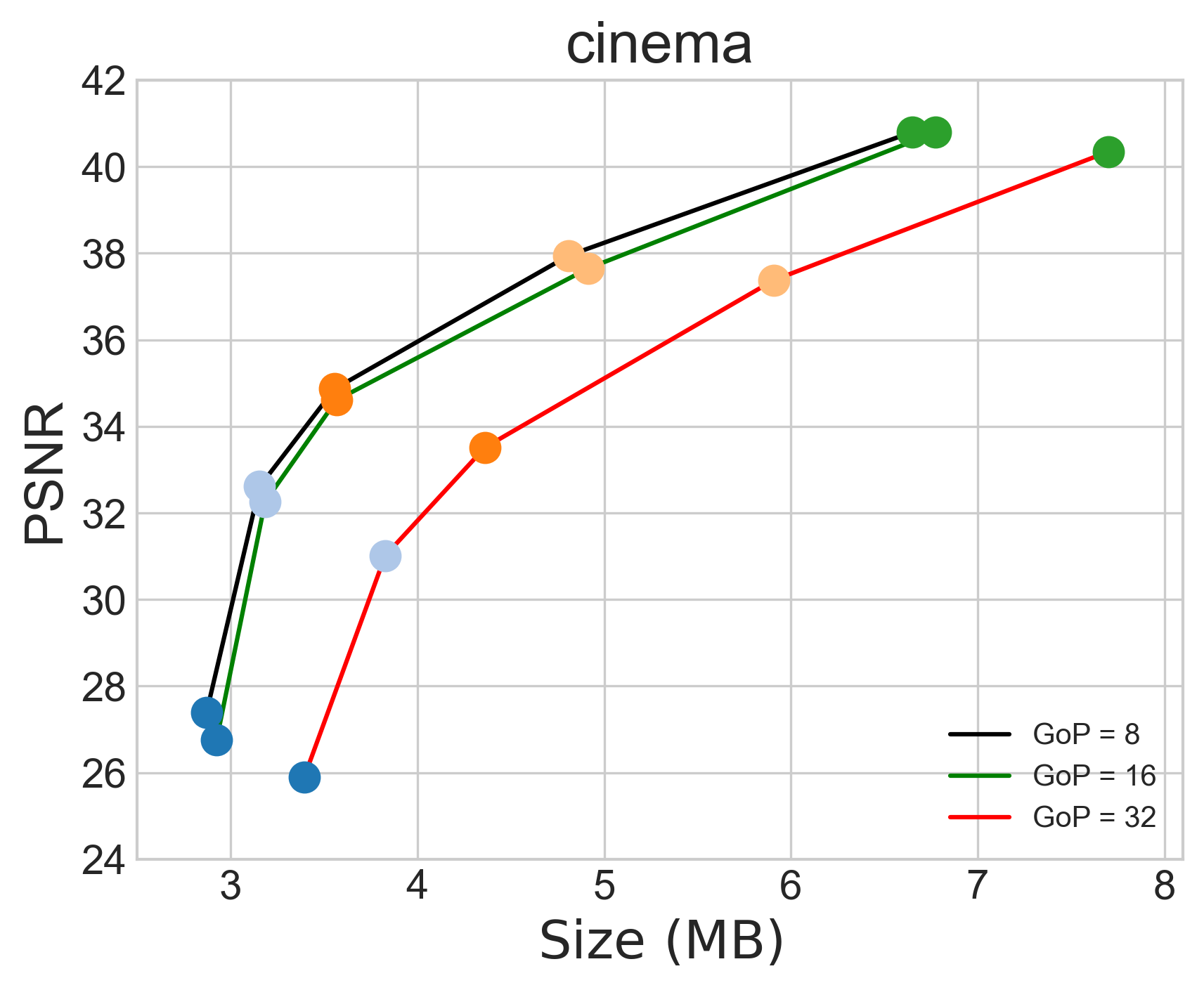}\end{minipage}}\\
\subfloat[Frame-wise PSNR at R03.]{\begin{minipage}{0.45\textwidth}\includegraphics[width=\linewidth]{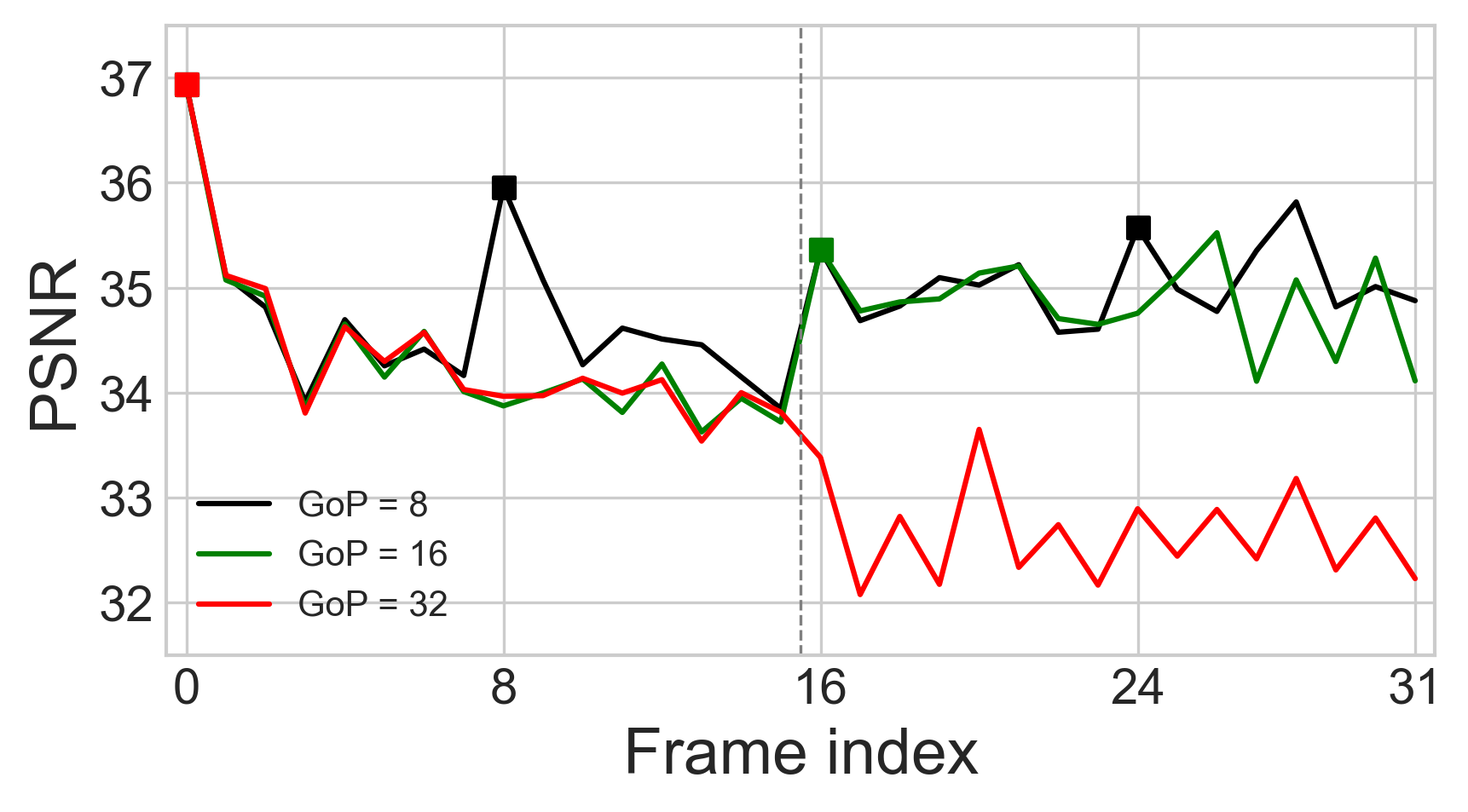}\end{minipage}}\\
\caption{(a): Performance of different GS GoP sizes on the 32-frame semi-tracked ``cinema'' sequence; (b): frame-wise PSNR at R03, squares mark I-frames, the dashed line separates frames 0--15 and 16--31.}
\label{fig:ab_gop}
\end{figure}

\begin{table}[h]
\caption{GoP-size comparison on the 32-frame ``cinema'' sequence.  RD changes use all five operating points with GoP8 as the reference. Temporal diagnostics are measured at R03.}
\label{tab:ab_gop}
\centering
\resizebox{\linewidth}{!}{%
\begin{tabular}{|l|ccc|}
\hline
Metric & GoP8 & GoP16 & GoP32 \\ \hline
RD gain (\%) & 0.00 & +3.91 & +28.04 \\ \hline
PSNR, frames 0--15 (dB) & 34.70 & 34.35 & 34.37 \\ \hline
PSNR, frames 16--31 (dB) & 35.04 & 34.87 & 32.66 \\ \hline
PCA dim., frames 16--31 & 24 & 24 & 21 \\ \hline
Proj. error, frames 16--31 (\%) & 0.25 & 0.25 & 1.02 \\ \hline
\end{tabular}%
}
\vspace{0.3em}
\begin{minipage}{\linewidth}
\end{minipage}
\end{table}

Fig.~\ref{fig:ab_gop}(a) and Table~\ref{tab:ab_gop} show a small RD degradation for GoP16 and a larger degradation for GoP32, with loss of $3.91\%$ and $28.04\%$ relative to GoP8. At R03, GoP32 increases size from 3.557 to 4.362 MB/frame while reducing mean PSNR from 34.87 to 33.51 dB relative to GoP8. The quality gap is concentrated in the second half of the sequence. In Fig.~\ref{fig:ab_gop}(b), GoP16 and GoP32 have nearly identical average PSNR over frames 0--15 (34.35 and 34.37 dB), but GoP32 is 2.21 dB lower over frames 16--31. Refreshing the I-frame at frame016 in GoP8/16 also updates the PCA latent dimension from 21 to 24, whereas GoP32 keeps the 21-dimensional frame000 representation. We also report the projection error of EA-SHDR in Table~\ref{tab:ab_gop}, i.e., $\|X_c-\widehat{X}_c\|_F^2/\|X_c\|_F^2$, where $X_c$ is centered using the group's I-frame mean.  Consequently, the second-half normalized projection error is 1.02\% for GoP32 and approximately 0.25\% for GoP8/16. The 0.99 energy threshold selects the dimension only at the I-frame and does not guarantee the same retained energy for subsequent P-frames. This difference includes both basis refresh and dimension adaptation. These observations are consistent with poorer matching to the distant I-frame anchor and reduced compressibility that local P-MiniPLAS refinement cannot fully recover. Therefore, periodic I-frame refresh maintains more stable RD performance and supports the default GoP8 setting.

\section{Conclusion}\label{sec:conclusion}
This paper presents a comprehensive and effective 3D GS sequence compression framework via the video codec GSCV++. GSCV++ optimizes the end-to-end video-based 3D GS compression pipeline with EA-SHDR, P-MiniPLAS, and CA-GC. It reports SOTA performance in both intra- and inter-prediction modes, as well as faster GS 2D map generation and coding speeds. Based on the experimental results, we demonstrate the robustness, compatibility, and alignment with the requirements of standardization for lightweight 3D GS compression.

\textbf{Limitations:} The GS geometry coordinate LSB map compression is less compressible for standard video codecs. Additionally, using energy solely to control the latent dimension of color SH AC cannot achieve an optimized RD balance for some GS sequences. Moreover, PLAS images report different characteristics compared with natural images; even the slightest roughness will limit the performance of the encoder. Some learning-based methods, such as Sandwich Network \cite{yu2026codec}, might further improve the performance of standardized video codecs.





\bibliographystyle{IEEEtran}
\bibliography{bib}


 




\begin{IEEEbiography}
[{\includegraphics[width=1in,height=1.25in,clip,keepaspectratio]{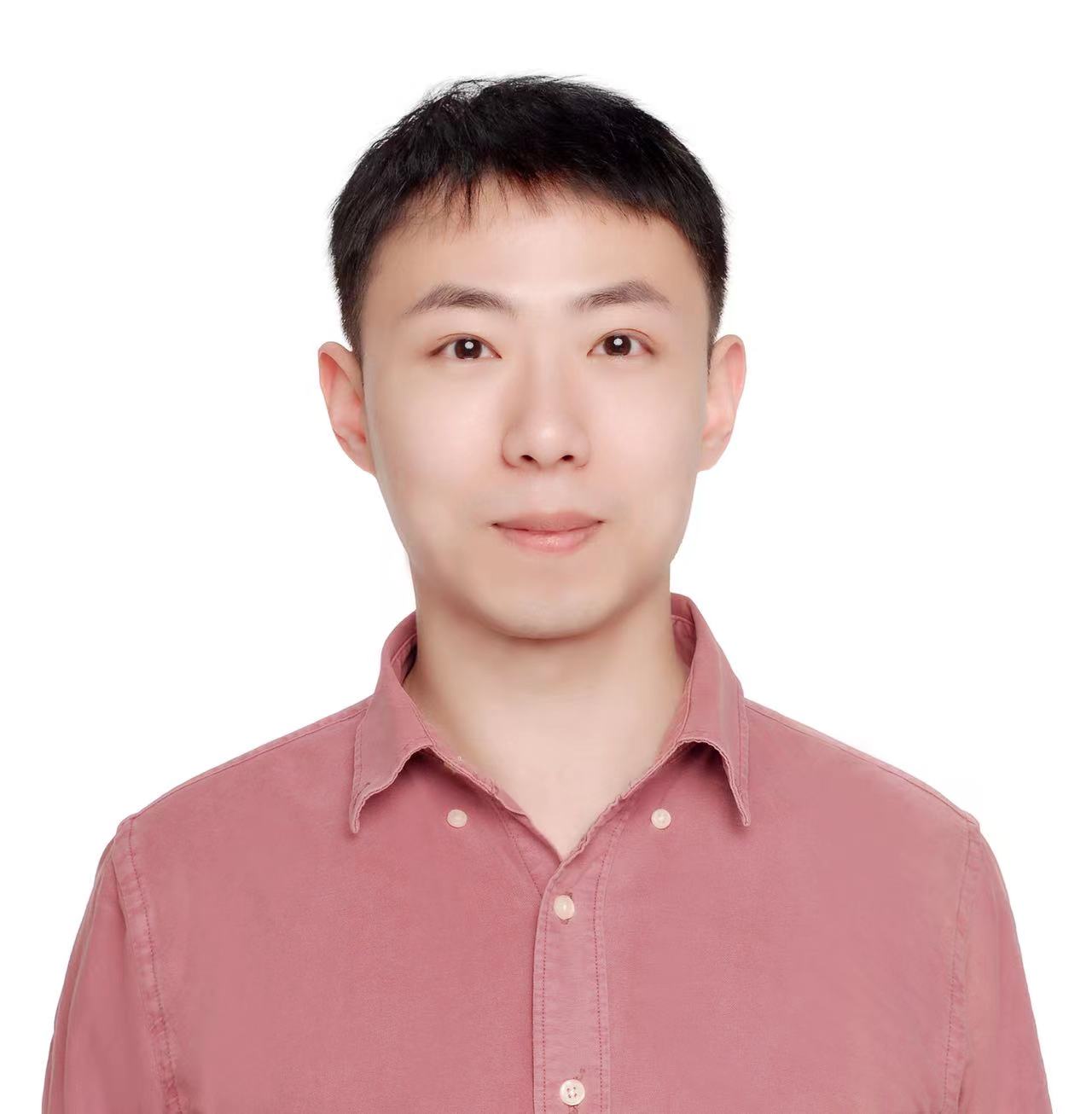}}]{Qi Yang} received the B.S. degree in communication engineering from Xidian University, Xi'an, China, in 2017, and Ph.D degree in information and communication engineering at Shanghai Jiao Tong University, Shanghai, China, 2022. He worked as a researcher in Tencent MediaLab from 2022 to 2024, and worked as a research associate at the University of Missouri–Kansas City from 2024 to 2026. Now, he is a video coding researcher at ByteDance Inc., San Diego, US. He has published more than 35 conference and journal articles, including TPAMI, TIP, TVCG, TMM, TCSVT, ICML, CVPR, ICCV, IJCAI, ACM MM, etc. He is also an active member in standard organizations, including MPEG, AOMedia, and AVS. His research interests include video coding, 3D/4D GS compact generation and compression, 3D point cloud and mesh quality assessment. He received the Best Paper Award of ACM TOMM in 2026, and Best Paper Candidate of VCIP in 2024.
\end{IEEEbiography}

\begin{IEEEbiography}[{\includegraphics[width=1in,height=1.25in,clip,keepaspectratio]{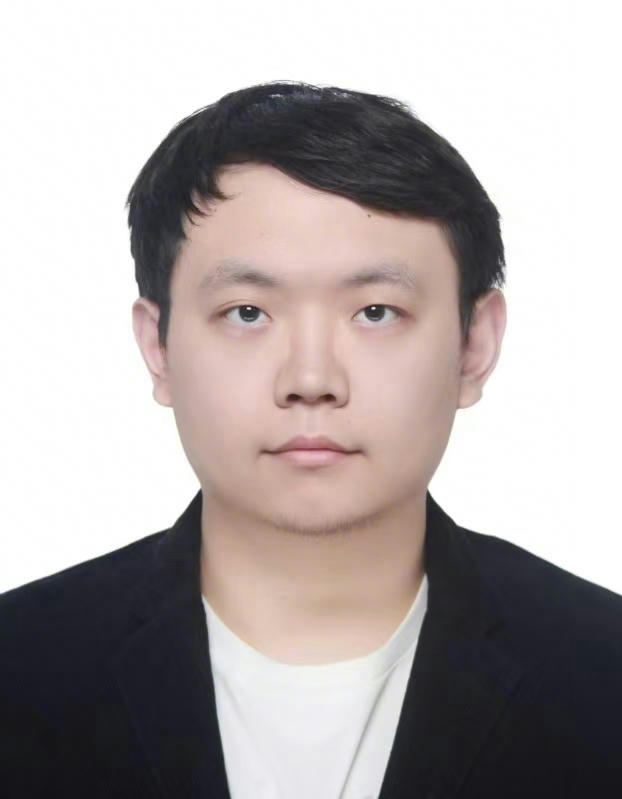}}]{Yingzhan Xu} received the B.S. and M.S. degrees in Communication Engineering from Shanghai Jiao Tong University, Shanghai, China, in 2018 and 2021, respectively. He is currently a video coding engineer in Bytedance Inc., Beijing. He was actively participating in the research of 3D graphics compression standards. His current research interests include point cloud compression and point cloud processing.  During the development of G-PCC and E-GPCC, he co-chaired the specification text editing and several core experiments.
\end{IEEEbiography}

\begin{IEEEbiography}[{\includegraphics[width=1in,height=1.25in,clip,keepaspectratio]{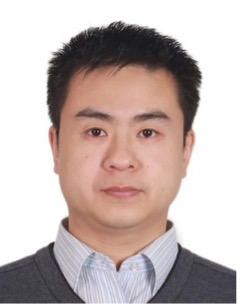}}]{Kai Zhang} (M’17, SM' 20) received the B.S. degree in computer science from Nankai University, Tianjin, China, in 2004. In 2011, he received the Ph.D. degree in computer science from the Institute of Computing Technology, Chinese Academy of Sciences, Beijing, China. From 2011 to 2012, he worked as a researcher in Tencent Inc. Beijing, China. From 2012 to 2016, he worked as a team manager in Mediatek Inc. Beijing, China, leading a research team to propose novel technologies to emerging video coding standards. From 2016 to 2018, he worked in Qualcomm Inc. San Diego, CA, still focusing on video coding standardization. Now, he is leading the standardization team in Bytedance Inc. San Diego, CA. Dr. Zhang’ research interests include video/image compression, coding, processing and communication, especially video coding standardization. From 2006, he has contributed more than 500 proposals to JVT, VCEG, JCT-VC, JCT-3V, JVET and AVS, covering many important aspects of major standards such as H.264/AVC, HEVC, 3D-HEVC, VVC and AVS-1,2,3. He has 800+ granted or pending U.S. patents applications. Most of these patents are essential to popular video coding standards. During the development of VVC, Dr. Zhang co-chaired several core experiments and branch of groups. Currently, Dr. Zhang serves as a coordinator of the reference software known as ECM in JVET, to explore video coding technologies beyond VVC. Dr. Zhang has co-authored 100+ papers and reviewed 80+ papers on top-tier journals/conferences. He is the AE of IEEE T-CSVT and IET-IP and a TPC member of DCC 2024~2027. He was an organizer of the Grand Challenge on Neural Network-based Video Coding in ISCAS 2022~2026. 
\end{IEEEbiography}

\begin{IEEEbiography}[{\includegraphics[width=1in,height=1.25in,clip,keepaspectratio]{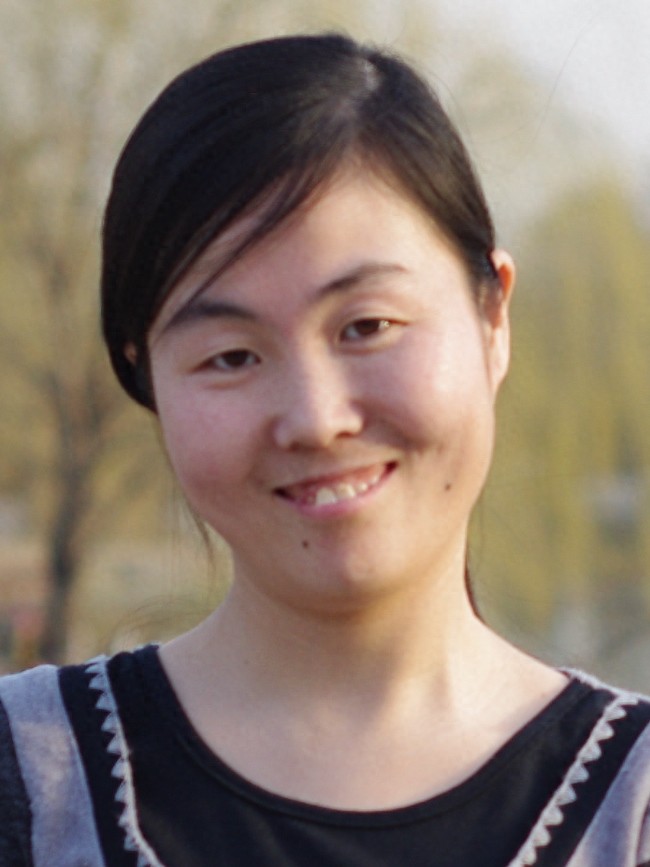}}]{Li Zhang}  (M’11–SM’20) received the Ph.D. degree in Computer Science from the Institute of Computing Technology, Chinese Academy of Sciences, in 2009. She currently leads the Multimedia Lab at ByteDance, San Diego, CA, USA. Prior to this, she held research positions at Qualcomm Inc. and Peking University.

Her research interests include image and video coding and AI-driven multimedia processing, understanding, and generation, covering both 2D and 3D representations. She has made significant contributions to international standards, including VVC (H.266), AVS, MPEG G-PCC, IEEE 1857, JPEG AI, and 3D extensions of HEVC (H.265). She has co-chaired multiple ad hoc groups and core experiments in standardization, served as an Editor for AVS, and was the Main Editor of the Software Test Model for 3DV. Her contributions have been recognized with multiple Certificates of Appreciation from the IEEE Standards Association and AVS. Her work has been widely adopted in international standards and industry deployments.

Dr. Zhang has authored over 220 technical publications and made extensive contributions to standardization, including more than 600 adopted proposals and over 1,000 granted U.S. patents. Her work has received several honors, including the 2026 IEEE Transactions on Circuits and Systems for Video Technology (T-CSVT) Best Paper Award, Best Paper Award at ISCAS 2022 (Visual Signal Processing and Communications track), Top 15 Best Paper at ICME 2025, and Top 10 Best Paper at PCS 2021. She has also led teams to first place in major international challenges, including CVPR CLIC, CVPR NTIRE, and ECCV AIM.

She served as an Associate Editor for IEEE Transactions on Circuits and Systems for Video Technology from 2022 to 2025 and has been an Associate Editor for IEEE Transactions on Image Processing since 2026. She has taken on key roles in major conferences, including Area Chair for AAAI 2027, ICME 2025, ICME 2026, Area Chair for the Multimedia Signal Processing Track at ICASSP 2026-2027, and Senior Program Committee Member for AAAI 2026. She has also contributed to the organization of leading conferences such as ACM Multimedia (ACMMM), ICIP, MMSP, and PCS, and actively organizes special sessions and grand challenges. She was a panelist at DCC 2025 and an industry keynote speaker at ICIP 2025.
\end{IEEEbiography}

\vfill

\end{document}